\def\year{2024}\relax
\documentclass[letterpaper]{article} 
\usepackage{aaai25}  
\usepackage{times}  
\usepackage{helvet}  
\usepackage{courier}  
\usepackage[hyphens]{url}  
\usepackage{graphicx} 
 \usepackage{multirow}
\usepackage{natbib}  
\usepackage{caption} 
\usepackage{color}
\usepackage{xcolor}
\DeclareCaptionStyle{ruled}{labelfont=normalfont,labelsep=colon,strut=off} 
\usepackage{algorithm}
\usepackage{algorithmic}
\usepackage{subfigure}
\usepackage{amsmath}
\usepackage{newfloat}
\usepackage{listings}
\floatstyle{ruled}
\newfloat{listing}{tb}{lst}{}
\floatname{listing}{Listing}
\title{Design Principle Transfer in Neural Architecture Search via Large Language Models}
\author{
    Xun Zhou$^1$, Xingyu Wu$^{1}\thanks{Corresponding author}$, Liang Feng$^{2*}$, Zhichao Lu$^3$, Kay Chen Tan$^1$
}
\affiliations{
    \textsuperscript{\rm 1}Department of Data Science and Artificial Intelligence, The Hong Kong Polytechnic University\\
    \textsuperscript{\rm 2}College of Computer Science, Chongqing University\\
    \textsuperscript{\rm 3}Department of Computer Science, City University of Hong Kong\\
    \{xingy.wu, kctan\}@polyu.edu.hk, xunzhou6-c@my.cityu.edu.hk, 
    liangf@cqu.edu.cn, luzhichaocn@gmail.com   
}

\usepackage{bibentry}
\begin{document}
\maketitle
\begin{abstract}
Transferable neural architecture search (TNAS) has been introduced to design efficient neural architectures for multiple tasks, to enhance the practical applicability of NAS in real-world scenarios. In TNAS, architectural knowledge accumulated in previous search processes is reused to warm up the architecture search for new tasks. However, existing TNAS methods still search in an extensive search space, necessitating the evaluation of numerous architectures. To overcome this challenge, this work proposes a novel transfer paradigm, i.e., design principle transfer. In this work, the linguistic description of various structural components' effects on architectural performance is termed design principles. They are learned from established architectures and then can be reused to reduce the search space by discarding unpromising architectures. Searching in the refined search space can boost both the search performance and efficiency for new NAS tasks. To this end, a large language model (LLM)-assisted design principle transfer (LAPT) framework is devised. In LAPT, LLM is applied to automatically reason the design principles from a set of given architectures, and then a principle adaptation method is applied to refine these principles progressively based on the new search results. Experimental results show that LAPT can beat the state-of-the-art TNAS methods on most tasks and achieve comparable performance on others. Code is available at https://github.com/milkmilk511/LAPT.

\end{abstract}

\section{Introduction}
Neural architecture search (NAS) has recently become a prominent research focus for the automatic construction of deep neural networks (DNNs)~\cite{zhou2021survey}. With the advancement of NAS methods, automatically generated DNNs surpass manually crafted ones across various deep learning tasks~\cite{zhou2021survey2}. Nonetheless, the majority of NAS methods concentrate on constructing DNN architectures for a single task. This implies that these methods must run repeatedly from scratch when networks for multiple tasks are required. This constraint impedes the practical utility of NAS in real-world settings.


To address this limitation, transferable NAS (TNAS) has been introduced~\cite{zhou2023towards,li2021meta, elsken2020meta,lu2021neural,huang2022arch, Hanxiao2018}. TNAS leverages architectural knowledge gained from prior search processes to expedite the architecture search for new tasks, effectively ``warming up" the search process. For instance, previously discovered architectures can inform and guide the search toward more promising configurations for a new task~\cite{zhou2023towards,li2021meta, elsken2020meta}. Similarly, performance predictors built during earlier NAS tasks can be reused to reduce the computational cost of evaluating new architectures~\cite{lu2021neural,huang2022arch}. Nevertheless, these approaches still operate within vast search spaces (e.g., 10$^{18}$ candidate architectures in DARTs~\cite{Hanxiao2018}), necessitating the evaluation of a prohibitively large number of architectures and thus resulting in an inefficient search process. Additionally, the current TNAS methods often intertwine the utilization of transferred knowledge with specific search techniques, limiting their adaptability to different NAS methodologies.


This paper introduces a novel paradigm called design principle transfer, aimed at overcoming these challenges. In this approach, design principles—expressed as linguistic descriptions of how certain structural components (e.g., layers or connections) affect the performance of architectures—are first extracted from architectures developed for previous tasks. These principles are then applied to prune the search space by eliminating architectures with less critical components, resulting in an optimized subspace for the new task. By searching within this refined subspace, the proposed approach aims to significantly enhance both the efficiency and performance of NAS for new tasks. Importantly, because knowledge reuse is decoupled from the architecture search itself, this paradigm is compatible with most NAS methods.


Despite its promise, the practical implementation of design principle transfer presents several challenges. Firstly, the complexity and diversity of DNN architectures make it difficult to distill general design principles. Current methods often require specialized tools to map architectures into a shared latent space, followed by expert analysis to extract underlying design rules~\cite{yuan2022visual}, which reduces the level of automation. Secondly, the process of learning these principles is resource-intensive, requiring a vast number of labeled architectures. For instance, \citeauthor{radosavovic2020designing}~(\citeyear{radosavovic2020designing}) trained over 500 architectures to investigate the relationship between architectural width and performance, incurring costs that exceed those of most NAS efforts. Additionally, the high-level abstraction of knowledge in natural language complicates its translation into actionable insights for architecture design.

With the emergence of pre-trained Large Language Models (LLMs)~\cite{wu2024evolutionary,liu2024evolution}, LLMs offer a promising solution to address the aforementioned challenges. By representing architectures in programming languages, the task of learning design principles can be framed as a language-inductive learning problem, a domain where LLMs have demonstrated proficiency~\cite{imani2023mathprompter}. Therefore, leveraging LLMs as reasoning agents for automatic design principle learning is a logical step. Given their pre-training on vast knowledge, in-context learning can be employed to tackle this task, thereby mitigating the constraints posed by the number of architecture samples. Furthermore, owing to their contextual awareness, LLMs can automatically translate design principles into actionable architectural knowledge for NAS methods.

Keeping the above in mind, this work proposes an LLM-assisted design principle transfer (LAPT) framework. The LAPT framework employs a pre-trained LLM to learn design principles from well-established neural architectures represented in programming codes. These principles are then translated into constraints that refine the predefined search space, optimizing it for new NAS tasks. However, due to domain shifts, the refined subspace may not always be optimal for every task. To address this, we introduce a principle adaptation method that refines the design principles based on the architectures found for the target task, thereby building the search space to the specific requirements of this task. The main contributions of this paper are summarized as follows:

\begin{itemize}
\item To the best of our knowledge, this work is the first research for the design principle transfer. This novel transfer paradigm aims to build a refined search space for new NAS tasks, leading to the improvement of search performance and efficiency.


\item An LLM-assisted framework is proposed to implement the design principle transfer across different NAS tasks, which offers at least three advantages: (i) Learning of the general design principles based on LLMs; (ii) Task-specific principle adaptation against domain shit; (iii) Improved interpretability of search space refinement.

\item Extensive experiments across various search spaces and tasks demonstrate the effectiveness of LAPT. Even when using standard NAS methods, searching within the refined search space leads to state-of-the-art (SOTA) results, highlighting the potential of design principle transfer as a promising research direction in NAS.
\end{itemize}

\section{Related Work}
This section reviews the most closely related studies to our work, namely TNAS and the utilization of LLMs in NAS. Additional related works can be found in \textbf{Appendix I}.

\subsection{TNAS}
To reduce the computational cost, transfer learning has been incorporated into NAS, giving rise to TNAS~\cite{huang2022arch,lee2021rapid,lu2020multiobjective,lu2021neural,elsken2020meta,zhou2023towards}. In TNAS, the knowledge gained during the architecture search for one task is applied to aid in the architecture design for others. Depending on the types of transferred knowledge, TNAS can be classified into two categories. One category transfers already found architectures from previous tasks to directly build architectures for new tasks. For instance, Lu et al. \cite{lu2020multiobjective} design architectures for CIFAR-100 and use them to classify images from ImageNet directly. However, the generalization performance of these methods has been criticized. To solve this drawback, multitasking evolutionary NAS (MTNAS)~\cite{zhou2023towards} applies these high-performing architectures to guide the architecture search for new tasks instead of solving them directly.

The second category reuses the NAS model built in previous tasks to guide the architecture search for a new task. For example, \citeauthor{lu2021neural}~(\citeyear{lu2021neural}) leverage the supernet that has been initialized on ImageNet to improve the search efficiency of evolutionary NAS (ENAS) on new image classification tasks. In another related work~\cite{elsken2020meta}, a meta-architecture is acquired from previous tasks, and this architecture can be adapted to a new task through just a few gradient-based steps. \citeauthor{huang2022arch}~(\citeyear{huang2022arch}) fine-tune the performance predictor developed for the Jigsaw task and use it to reduce the search cost on other computer vision tasks. However, these methods still search in a very large search space, consuming a lot of computing resources for obtaining high-quality architectures. In this work, a new paradigm of TNAS is proposed, i.e., design principle transfer. In this method, general design principles are summarized from established architectures and further reused to reduce the search space for new NAS tasks.  

 \begin{figure*}[t]
    \centering
    \includegraphics[width=1\linewidth]{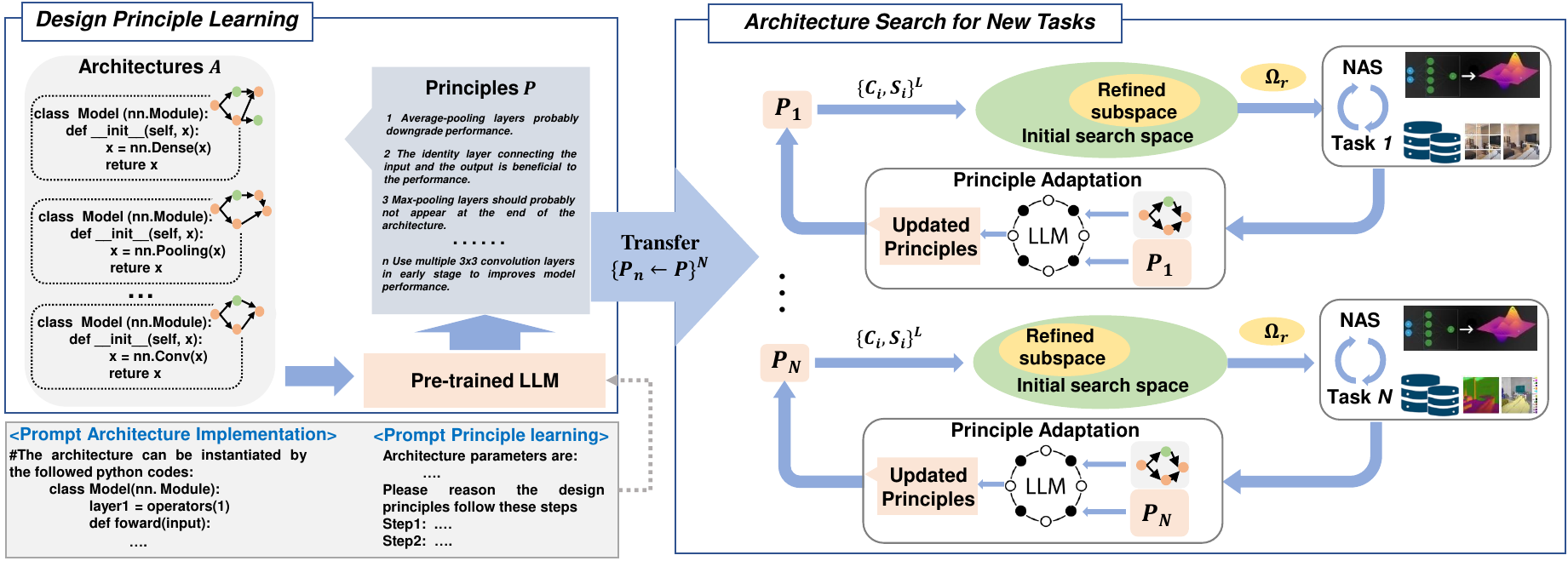}
    \caption{Overview of the proposed LAPT. This framework consists of two stages. In the learning stage of design principles, LLM is driven by specific prompts to learn general design principles from a set of architectures. In the architecture search stage, the learned principles are applied to initialize the search space for each new task. Then, architectures found in the refined search space are used to update these principles, aiming to build the task-specific search space.}
    \label{overview}
\end{figure*}
\subsection{LLMs For NAS}
Most existing NAS methods still follow a semi-automated process, where designs for search space, search method, and performance evaluation system rely on expert knowledge~\cite{chen2024evoprompting}. To improve the automation degree of architecture design, LLMs have been incorporated into NAS, because of their robust capabilities in solving various domain-specific tasks. For instance, leveraging its powerful generative capabilities, GPT-4 has been utilized to generate convolution neural networks (CNNs)~\cite{chen2024evoprompting,zheng2023can,nasir2023llmatic,qin2024fl} and graph neural networks (GNNs)~\cite{wang2023graph,dong2023heterogeneous}. When compared to many mainstream NAS methods, these LLM-based methods can achieve comparable performance. Moreover, LLMs can serve as a performance predictor within NAS to accelerate the search process~\cite{zhang2023automl,jawahar2023llm, chen2024large}. In contrast to performance predictors developed using machine learning methods or DNNs, LLMs can achieve comparable performance with fewer training samples. Contrary to existing works focusing on the refinement of the search method and the performance evaluation system, our work applies LLMs to refine search space, further extending the application of LLMs in NAS research.

\section{Proposed Method}
In this section, we initially define the architecture search problem and introduce a design principle transfer framework to effectively address it. Subsequently, details of these key components within this framework are presented including design principle learning and principle adaptation.
\subsection{Problem formulation}
Given a task $\mathcal{T}$ associated with labeled dataset $\mathcal{D}$, we consider searching an architecture $a^*$ from a predefined search space $\Omega$ to achieve the best accuracy (termed $ACC.$) on this task. Different from existing NAS works that directly search in the entire search space, we only explore a promising subspace $\Omega_r$ for the efficient architecture search. This NAS problem can be formulated as follows:
\begin{equation}
\begin{split}
    a^* = &\mathop{\arg\max}\limits_{a\in\Omega_r}ACC.(a,~\mathcal{D})\\
    &s.t.~\Omega_r\subset\Omega, 
\end{split}
\end{equation}
where the key to solving equation (1) is building the optimal subspace $\Omega_r$ of $\Omega$. In this work, $\Omega$ consists of a set of available architectures for the target task $\mathcal{T}$. For each architecture $A$ in $\Omega$, it follows a feed-forward structure with no more than $L$ layers. Each layer $l_i $ is associated with some operators (such as pooling and convolution) to deal with the received feature information, where $\mathcal{C}_i$ denotes the candidate operators for the $i$th layer. Additionally, the feed-forward structure makes $l_i$ only receive information from its previous layers $\mathcal{S}_i=\{l_1,\ldots, l_{i-1}\}$, where $s_i$ denotes the information sources of the $i$th layer. Thus, an architecture $A$ can be parameterized as $A={\cup}_{i=1}^L\{l_i,~s_i\}$, and $\Omega$ can be formulated as equation (2): 
\begin{equation}
\Omega= \{A|~l_i\in\mathcal{C}_i,~s_i\subset \mathcal{S}_i, i\in\{1,\ldots,~L\}\}.
\end{equation}
To extract the optimized subspace from $\Omega$, design principles that describe the influence of various operators and connections for the architecture performance can be used to discard the unimportant operators and information sources of each layer. In this way, a subspace for each layer, i.e., refined candidate operators $\mathcal{C}^r_i\subset\mathcal{C}_i$ and refined candidate information sources $\mathcal{S}_i^r\subset\mathcal{S}_i$, can be built. The refined search space $\Omega_r$ is in equation (3): 
\begin{equation}
\Omega_r= \{a|~l_i\in\mathcal{C}^r_i,~~s_i\subset \mathcal{S}_i^r,~a\in\Omega\}.
\end{equation}
Compared with $\Omega$, $\Omega_r$ has a higher proportion of well-performing architectures, searching in this optimized search space is more efficient to reach the architecture with good performance. The quality analysis of $\Omega$ and $\Omega_r$ is presented in \textbf{Appendix III}.

\subsection{Framework} An overview of LAPT is presented in Figure~\ref{overview} and Algorithm 1. This framework consists of two stages, i.e., design principle learning and the architecture search for new tasks. In the design principle learning stage, a set of well-performing architectures $\mathcal{A}$ is collected from a pre-defined search space $\Omega$ (line 1 in Algorithm 1). Then, a prompt is devised to help a pre-trained LLM reason general design principles $P$ from $\mathcal{A}$ (line 2 in Algorithm 1). These learned principles $P$ are transferred as the initial design principles $\{P_{n}\}_{n=1}^N$ to help solve $N$ new NAS tasks $\{\mathcal{T}_n\}_{n=1}^N$ (lines 3-10 in Algorithm 1). Specifically, for the task $\mathcal{T}_n$, $P_n$ is translated to the subsets $\{\mathcal{C}^r_i,~\mathcal{S}^r_i\}_i^L$ by the pre-trained LLM, leading to the generation of a refined search space $\Omega_{r}$. Then, a NAS method is applied to search for the promising architectures $\mathcal{B}_{n}$ from $\Omega_{r}$ for $\mathcal{T}_n$. These found architectures are applied to adapt $P_n$ to $\mathcal{T}_n$. These steps related to principle adaptation and architecture search are repeated until reaching a predefined stopping criterion. Finally, the best architecture found in the search process is used to solve $\mathcal{T}_n$.
\begin{algorithm}[t]
\caption{Framework of the proposed LAPT}
\label{alg:algorithm}
\textbf{Input}: Search space $\Omega$, new tasks $\{\mathcal{T}_n\}_{n=1}^N$, a pre-trained LLM, a NAS method, iterations $G$.\\
\textbf{Output}: Architecture $\{a_n^*\}_{n=1}^N$.
\begin{algorithmic}[1] 
\STATE Build an archive $\mathcal{A}$ consisting of performing architectures from $Omega$;
\STATE $\{P_{n}\}_{n=1}^N\leftarrow$ Prompt the LLM to learn design principles $P$ from $\mathcal{A}$ and set it as initial principles for each task;
\FOR{$n$ from 1 to $N$}
\STATE $Base\leftarrow0$;
\FOR{$g$ from 1 to $G$}
\STATE $\Omega_{r}\leftarrow$ LLM translate $P_{n}$ to a set of subset $\{\mathcal{C}^r_i,~\mathcal{S}^r_i\}_i^L$ of $\Omega$ and build subspace follow (3);
\STATE $\mathcal{B}_{n}\leftarrow$ Search for architectures for the task $\mathcal{T}_{n}$ from $\Omega_{r}$ by the given NAS method;
\STATE $Best\leftarrow$ Evaluate architectures $\mathcal{B}_{n}$ on $\mathcal{T}_{n}$ and record the performance of the best one;
\STATE $P_{n},~Base\leftarrow$ Adapt $P_n$ based on $\mathcal{B}_{n}$, $Base$, $Best$ and return the updated principles; \#Alg. 2
\ENDFOR
\ENDFOR
\STATE Return the best architecture $\{a_n^*\}_{n=1}^N$ found in the search process.
\end{algorithmic}
\end{algorithm}

\subsection{Design principle learning}
In this part, a prompt is designed to guide the LLM to learn the design principles from the given architectures. The prompt consists of two parts, i.e., architecture implementation and learning guidelines.

\begin{figure}[t]
    \centering
    \includegraphics[width=1\linewidth]{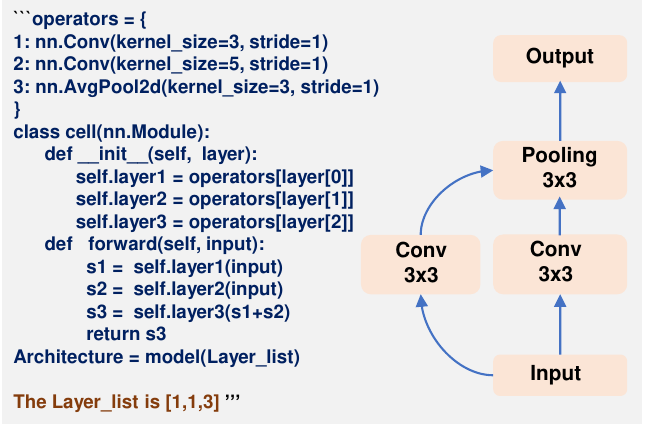}
    \caption{The Python code-based prompt can help LLMs gain awareness of the right CNN architecture from its architectural parameters which are represented as ``Layer\_list".}
    \label{prompt_imp}
\end{figure}

Firstly, the pre-trained LLM benefits from exposure to a wide array of programming languages, allowing it to gain awareness of the neural architecture from source codes~\cite{zheng2023can}. Nevertheless, due to the token limitation, it becomes infeasible to feed all architecture source codes directly into the LLM. To tackle this issue, Python classes that can instantiate an architecture from its architectural parameters, i.e., ${\cup}_{i=1}^L\{l_i,~s_i\}$, are set as prompts. This approach enables LLMs to assimilate knowledge about these neural architectures solely through a few architecture parameters. A simple example of the Python code-based prompt is shown in Figure.~\ref{prompt_imp}.

Secondly, instructing LLMs to reason the general design principles from such architectures is not trivial, given the complex and diverse DNN architectures. To address this issue, drawing inspiration from the effective utilization of the ``chain of thought" method in LLMs, we steer the LLM towards a step-by-step reasoning process as follows:
\begin{itemize}
\item \textbf{Step1:} input architectural parameters of the given architectures into the LLM;

\item \textbf{Step2:} prompt LLM identifying common patterns within these architectures;

\item \textbf{Step3:} summarize the design principle behind these common patterns.
\end{itemize}

More Details of this prompt can be found in \textbf{Appendix II}.
\subsection{Principle transfer and adaptation}
In the architecture search stage, principles $P$ learned in the last stage are used to help build the refined search space $\Omega_r$. Specifically, we prompt the pre-trained LLM to translate $P$ into the most available candidate operators $\mathcal{C}_i$ and information sources $\mathcal{S}_i$ for each layer. In this way, $\Omega_r$ can be built following equation (3). Then, the NAS method is used to search architectures $\mathcal{B}_{n}$ from $\Omega_r$, aiming to solve the problem shown in equation (1).

However, due to domain shift where the architecture performs differently on different tasks, $\Omega_r$ may not be optimal for all the tasks. To alleviate the potential negative effects of domain shift, this work adapts $P$ to the target task based on the newly found architectures. As a result, a task-specific subspace can be built progressively. Specifically, the NAS method is applied to search for architecture from $\Omega_r$ in an iterative way. In each iteration, if the better-performing architectures are found, LLM is prompted to update $P_n$ based on these architectures; otherwise, LLM is required to describe the effects of other available candidate operators and information sources that are not in $P_n$, promoting the exploration for other promising regions in $\Omega$. Details of this adaption strategy can be seen in Algorithm 2.
\begin{algorithm}[tb]
\caption{Principle Adaption}
\label{alg:algorithm}
\textbf{Input}: design principles $P_n$, found architectures $\mathcal{B}_{n}$, $Base$, $Best$, \# of selection architectures $r$.\\
\textbf{Output}: updated $P_n$.
\begin{algorithmic}[1] 
\IF{$Base \leq Best$}
\STATE $P_{n}\leftarrow$ Prompt LLM to update $P_{n}$ from 
 the top $r$ architectures of $\mathcal{B}_{n}$;
\STATE $Base\leftarrow Best$;
\ELSE
\STATE $P_{n}\leftarrow$ Prompt LLM to describe effects of other operators and connections that are not in current $P_n$;
\ENDIF
\STATE Return $P_n$, $Base$.
\end{algorithmic}
\end{algorithm}


\section{Experiments}
To demonstrate the efficacy of the proposed LAPT, we perform architecture searches for diverse tasks utilizing the design principles learned from established architectures. Subsequently, we compare its search performance with other well-known NAS and TNAS methods. This section commences with a description of the search spaces and experimental setup, followed by the presentation of comparisons.
\subsection{Search space}
This work tests the proposed TNAS method on three search spaces, i.e., NAS-bench-201 (NAS201)~\cite{dong2020bench}, TransNAS-Bench-101 (Trans101)~\cite{duan2021transnas}, and DARTs~\cite{Hanxiao2018}.

\textbf{NAS201.} The architecture in NAS201 consists of the repeated cells. Each cell consists of 6 layers and there are five candidate operators for each layer. These layers are connected following a predefined pattern leading to 15K architectures in this search space.

\textbf{Trans101.} Architectures in Trans101 follow the same cell-based structure in NAS201, but there are only 4 candidate operators for each layer. Thus, only 4K candidate architectures are included in the space.

\textbf{DARTs.} Architectures in this search space also follow the cell-based structure. Particularly, each architecture consists of stacking two types of cells, i.e., the normal cell and the reduction cell. For each cell, there are 2 input layers and 4 ordered stages. Each stage includes two layers, and each layer has 8 candidate operators to deal with information from previous stages or input layers, resulting in $\prod_{k=1}^4\frac{k(k+1)}{2}8^2\approx 10^9$ candidate cell structures. Since we jointly learn both normal and reduction cells, the total number of architectures is $(10^9)^2$.
\begin{table}[t]
  \centering
  \caption{Hyper-parameters settings}
  \label{notations}
  \scriptsize
  \begin{tabular}{lllll}
    \hline
    \hline
    {\bf Stage}&{\bf Parameter}&{\bf NAS201}&{\bf Trans101}&{\bf DARTs}\\
    \hline
    \vspace{0mm}
   {Learning}&\# of samples&50&50&100\\
    \hline
    \multirow{2}*{Adaption}&$r$&5&15&50\\
    {}&\# of iterations&3&4&2\\
    \hline
    {}&population size&5&10&20\\
    {}&\# of generations&1&1&10\\
    {REA}&tournament size&2&5&2\\
    {}&crossover probability&-&-&0.5\\
    {}&mutation probability&1&1&0.5\\
    \hline
    {Supernet}&\# of epochs&-&-&30\\
    \hline
    \hline
  \end{tabular}
  \label{hyp}
\end{table}

\subsection{Experiment Setup}
LAPT is employed to search for task-specific architectures within the three distinct search spaces mentioned above. Given that this framework is adaptable to various NAS methods, the vanilla NAS method \textbf{Regular EA (REA)}~\cite{Real2019} is utilized in these experiments to showcase the efficacy of the design principle transfer. Details of hyper-parameters are in Table~\ref{hyp} and implementations are as follows:

\textbf{Experiments on NAS201.} In these experiments, we collect 50 top-performing architectures on ImageNet from NAS201 as the input for LLM. Then, learned principles are used to constrain the range of available operators on each layer, leading to a refined search space of NAS201. We separately search for top-performing architectures (i.e., Top 0.1\%) on CIFAR-10 and CIFAR-100 from the refined search space and record the time cost.

\textbf{Experiments on Trans101.} In these experiments, we collect 50 top-performing architectures on Jigsaw from Trans101 as the input for LLM. Then, learned principles are used to constrain the range of available operators on each layer, leading to a refined search space of Trans101. We search for architectures from the refined search space to solve 6 different computer vision tasks i.e., Object Classification (Obj), Scene Classification (SC), Room Layout (Roo), AutoEncoder (Auto), Surface Normal (Nor), Semantic Segmentation (Seg), respectively.

\textbf{Experiments on DARTs.} In these experiments, we randomly select 5000 architectures from DARTs and evaluate them on CIFAR-100. To reduce the evaluation cost, a supernet that covers all the architectures in DARTs is pre-trained on CIFAR-100 and each architecture can inherit weights from it for evaluation. The top 100 architectures in the selected architectures are the input of LLM. Then, learned principles are used to constrain the number of candidate operators on each layer and its candidate information sources, leading to a refined search space of DARTs. We search for individual architecture from the refined search space to solve image classification tasks on CIFAR-10 and ImageNet. To accelerate the search process, a supernet that covers the refined search space is built and then pre-trained on the target task. As a result, each architecture can inherit weights from it for evaluation. 

In these experiments, GPT-4 is used as the pre-trained
LLM for design principle learning and adaptation. Related
prompts can be seen in \textbf{Appendix II}. In addition, the effectiveness of different LLMs is shown in \textbf{Appendix V}.

\begin{table}[h]
  \centering
  \caption{Comparison with TNAS and NAS methods on NAS201.We present the test accuracy achieved by the found architectures on two unseen datasets. Additionally, we provide the number of neural architectures (\textbf{Archs.}) that are trained in the search. We run LAPT 20 times with different random seeds and average values are reported.}
  \scriptsize
  \begin{tabular}{lllll}
    \hline
    \hline
    \multirow{2}*{\bf Method}&\multicolumn{2}{c}{CIFAR-10}&\multicolumn{2}{c}{CIFAR-100}\\
    {}&{\bf Acc.(\%)}&{\bf Archs.}&{\bf Acc.}&{\bf Archs.}\\
    \hline
    \vspace{0mm}
    REA~\cite{Real2019}&93.92&$>$500&71.84&$>$500\\
    HEBO~\cite{cowen2022hebo}&94.34&100&72.62&100\\
    \hline
    \vspace{0mm}
   MetaD2A~\cite{lee2021rapid}&94.37&100&73.34&100\\
   TNAS-BO~\cite{shala2022transfer}&94.37&29&73.51&59\\
    \hline
    \vspace{0mm}
    \textbf{LAPT-REA}&\textbf{94.36}&\textbf{4.9}&\textbf{73.45}&\textbf{8.6}\\
    \hline
    \hline
  \end{tabular}
  \label{NAS201}
\end{table}

\begin{table*}[t]
  \centering
  \caption{Comparison results under same time cost on Trans101 (For Metric: $\uparrow$ indicates higher is better, $\downarrow$ indicates lower is better, \textbf{bold} indicates the best search results). LAPT runs 25 times with different random seeds.}
  \scriptsize
  \begin{tabular}{llllllll}
    \hline
    \hline
    {\bf Tasks}&{\bf Obj}&{\bf SC}&{\bf Roo}&{\bf Auto}&{\bf Nor}&{\bf Seg}&{\bf Total}\\
    {\bf Metric}&{\bf Acc$\uparrow$~~~~}&{\bf Acc$\uparrow$~~~~}&{\bf L2 loss$\downarrow$}&{\bf SSIM$\uparrow$~~~~}&{\bf SSIM$\uparrow$~~~~}&{\bf L2 MIoU$\uparrow$~~~~}&{\bf Ave. Rank$\downarrow$}\\
    \hline
    \vspace{0mm}
    REA~\cite{Real2019}&45.39\% &54.62\% &61.75 &\textbf{56.96}&57.22&25.52&38.50\\
    BONAS~\cite{shi2020bridging}&45.50\% &54.46\% &61.10 &56.73&57.46&25.32&34.31\\
    \hline
    \vspace{0mm}
    WeakNAS-T~\cite{wu2021stronger}&45.29\% &54.78\% &60.70&56.90 &57.19&25.41&35.73\\
    Arch-zero~\cite{huang2022arch}&45.64\% &54.80\% &60.21 &56.61&57.90&25.73&14.7\\
    Arch-Graph~\cite{huang2022arch}&45.81\% &\textbf{54.90\%} &\textbf{60.08} &56.58&\textbf{58.27}&25.69&\textbf{12.2}\\
    \hline
    \vspace{0mm}
    \textbf{LAPT-REA}&\textbf{45.96}\% &\textbf{54.89\%} &60.18 &56.52&57.69&\textbf{25.91}&\textbf{12.3}\\
    \hline
    \vspace{0mm}
    Global  Best~~&46.32\% &54.94\% &59.38 &57.72&59.62&26.27&1\\
    \hline
    \hline
  \end{tabular}
  \label{Trans101}
\end{table*}

\subsection{Comparison to NAS methods} 
\textbf{Comparison on NAS201.} We choose well-known NAS methods and TNAS methods as baselines: (1) the vanilla NAS method REA and the state-of-the-art NAS method HEBO~\cite{cowen2022hebo}; (2) the state-of-the-art TNAS methods include TNAS-BO~\cite{shala2022transfer} and MetaD2A~\cite{lee2021rapid}. In LAPT, REA is used for architecture search, and its results (LAPT-REA) are shown in Table~\ref{NAS201}.

 \begin{figure}[h]
    \centering
    \includegraphics[width=1\linewidth]{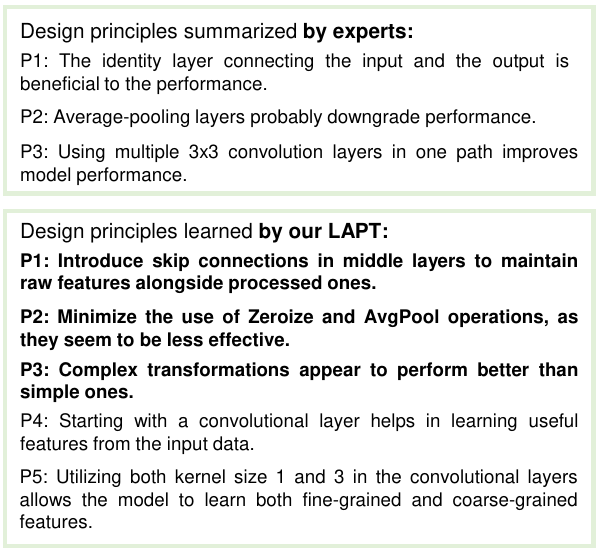}
    \caption{Design principles separately learned by LAPT and experts. Our LLM-based framework not only learns the identical principles (i.e., $P1-P3$) with experts but also can reason other valuable principles $P4$ and $P5$.}
    \label{principles_201}
\end{figure}

Compared with NAS methods, our proposed LAPT achieves better search performance and search efficiency by a large margin. Specifically, LAPT can reach the top architectures for CIFAR-10 and CIFAR-100 with 100x less time cost than REA and 10x less time cost than HEBO. 

Compared with TNAS methods, benefiting from the search space refinement, LAPT can achieve the SOTA performance. Specifically, all of these methods can reach the top architecture on CIFAR-10, but LAPT is about 25x faster than MetaD2A and 7x faster than TNAS-BO. On CIFAR-100, LAPT is about 12x faster than MetaD2A obviously, and is about 7x faster than TNAS-BO. 

We further show the general principles learned for the architectures in NAS201 in Figure.~\ref{principles_201} and the ones summarized by experts~\cite{yuan2022visual}. We can see that LAPT can learn the identical principles (i.e., P1-P3) with experts. Most importantly, LAPT can learn two other principles, i.e., P4 and P5. This study demonstrates that our LLM-based approach can automatically learn design principles and produce valuable and varied outcomes.

\textbf{Comparison on Transfer101.} We choose well-known NAS methods and TNAS methods as baselines: (1) the vanilla NAS method REA and the NAS method BONAS~\cite{shi2020bridging}; (2) the state-of-the-art TNAS methods include WeakNAS-T~\cite{wu2021stronger}, Arch-zero and Arch-Graph~\cite{huang2022arch}. All of these methods have the same search cost, i.e., the number of trained architectures is set to 50. Transferred architectural knowledge of all the TNAS is from the same computation task, i.e., Jigsaw.

From Table~\ref{Trans101}, our LAPT can achieve better search performance and search efficiency by a large margin than the NAS methods. Specifically, LAPT can beat BONAS on all the 6 tasks. LAPT outperforms REA on 5 out of 6 tasks and achieves the highest average mode rank. 

Compared with these TNAS methods, benefiting from the search space refinement, LAPT can achieve the SOTA results. Specifically, LAPT beats WeakNAS-T on most tasks (5 out of 6). Compared with the Arch-zero, LAPT achieves better performance on half of the tasks, i.e., Obj, Auto, and Seg obviously, and achieves comparable performance on other tasks. Compared with the SOTA method Arch-Graph, LAPT can beat it on tasks obj and Seg, and achieve similar performance on other tasks. The learned general principles
for Transfer101 are shown in \textbf{Appendix IV}.

\textbf{Comparison on DARTs.} In these experiments, LAPT is applied to search for architectures for CIFAR-10 and ImageNet from DARTs. Baselines include: (1) the basic CNNs designed by experts include VGG~\cite{simonyan2014very}, ResNet-110~\cite{he2016deep}, DenseNet-BC~\cite{huang2017densely}, and SENet~\cite{zagoruyko2016wide}; (2) the superior NAS methods which have improved state-of-the-art results at that time include SNAS~\cite{xie2018snas}, and DARTs~\cite{Hanxiao2018}; (3) SOTA NAS methods include GibbsNAS~\cite{xue2021rethinking}, MFENAS~\cite{yang2022accelerating}, and MASNAS~\cite{dong2022cell}. In LAPT, the vanilla NAS method REA is used for architecture search, and its search results (LAPT-REA) on CIFAR-10 are shown in Table~\ref{DARTsresult}.
\begin{table}[t]
  \centering
  \caption{Performance comparison on CIFAR-10 using the search space DARTs.}
  \scriptsize
  \begin{tabular}{lll}
    \hline
    \hline
    {\bf Architecture}&{\bf Test Acc.(\%)}&{\bf GPU Days}\\
    \hline
    \vspace{0mm}
    VGG~\cite{simonyan2014very}&93.35&-\\
    ResNet-110~\cite{he2016deep}&93.40&-\\
    DenseNet-BC~\cite{huang2017densely}&94.81&-\\
    SENet~\cite{zagoruyko2016wide}&95.38&-\\
    \hline
    \vspace{0mm}
    SNAS~\cite{xie2018snas}&97.15  &1.5\\
    DARTs~\cite{Hanxiao2018}&97.28 &4\\
    \hline
    \vspace{0mm}
    GibbsNAS~\cite{xue2021rethinking}&97.47  &0.5\\
    MSNAS~\cite{dong2022cell}&97.32&0.25\\
    MFENAS~\cite{yang2022accelerating}&97.61 &0.6\\
    \hline
    \vspace{0mm}
    \textbf{LAPT-REA}&\textbf{97.35} &\textbf{0.1}\\
    \hline
    \hline
  \end{tabular}
  \label{DARTsresult}
\end{table}

From Table~\ref{DARTsresult} we can observe that because of the search space optimization, LAPT can find the superior architectures more efficiently. Specifically, LAPT achieves superior performance in terms of test accuracy against these handcrafted CNNs including VGG-16, ResNet-110, DenseNet-BC, and SENet on CIFAR-10. Compared to superior NAS methods, LAPT can outperform them. To be specific, LAPT can beat SNAS using 15$\times$ fewer GPU days and cost 10$\times$ fewer GPU days to achieve better performance than DARTs. Although the test accuracy of LAPT is slightly worse than state-of-the-art methods, it has less time cost. Specifically, LAPT achieves comparable performance on CIFAR-10 with GibbsNAS, MSNAS, and MFENAS but costs 5$\times$, 2.5$\times$, and 6$\times$ fewer GPU days.

Search results on ImageNet are shown in Table~\ref{DARTsresult_IMAGE}. From this table, we can see that because of the search space refinement, LAPT can achieve similar search results with other NAS methods but with about 1.5-2$\times$ less search cost (2 GPU days), which is more time-efficient. The learned general principles
for DARTS are shown in \textbf{Appendix IV}.
\begin{table}[t]
  \centering
  \caption{Performance comparison on ImageNet using the search space DARTs.}
  \scriptsize
  \begin{tabular}{llll}
    \hline
    \hline
    {\bf Architecture}&{\bf Top1 Acc.(\%)}&{\bf Params.}&{\bf GPU Days}\\
    \hline
    \vspace{0mm}
    InceptionV1&69.8&6.6M&-\\
    MobileNet&70.6&4.2M&-\\
    ShuffleNet&73.7&5.0M&-\\
    \hline
    \vspace{0mm}
    PC-DARTs~\cite{xu2020pcdarts}&75.8&5.3M&3.8\\
    FairDARTs~\cite{chu2020fair}&75.6&4.3M&3.0\\
    Shapley-NAS~\cite{xiao2022shapley}&76.1&5.4M&4.2\\
    \hline
    \vspace{0mm}
    \textbf{LAPT-REA}&\textbf{75.1} &\textbf{4.6M} &\textbf{2.0}\\
    \hline
    \hline
  \end{tabular}
      \label{DARTsresult_IMAGE}
\end{table}

\subsection{Ablation study}
\begin{figure}[h]
\centering
\subfigure[Obj]{
    \begin{minipage}[t]{0.215\textwidth}
    \includegraphics[scale=0.275]{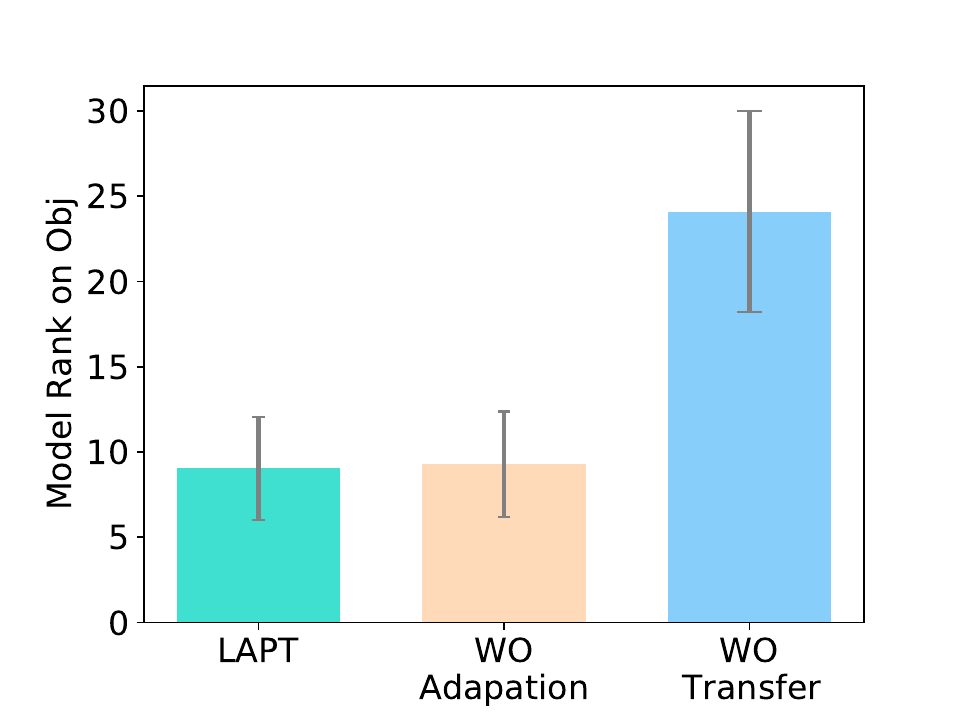}
    \end{minipage}
}
\subfigure[SC]{
    \begin{minipage}[t]{0.215\textwidth}
    \includegraphics[scale=0.275]{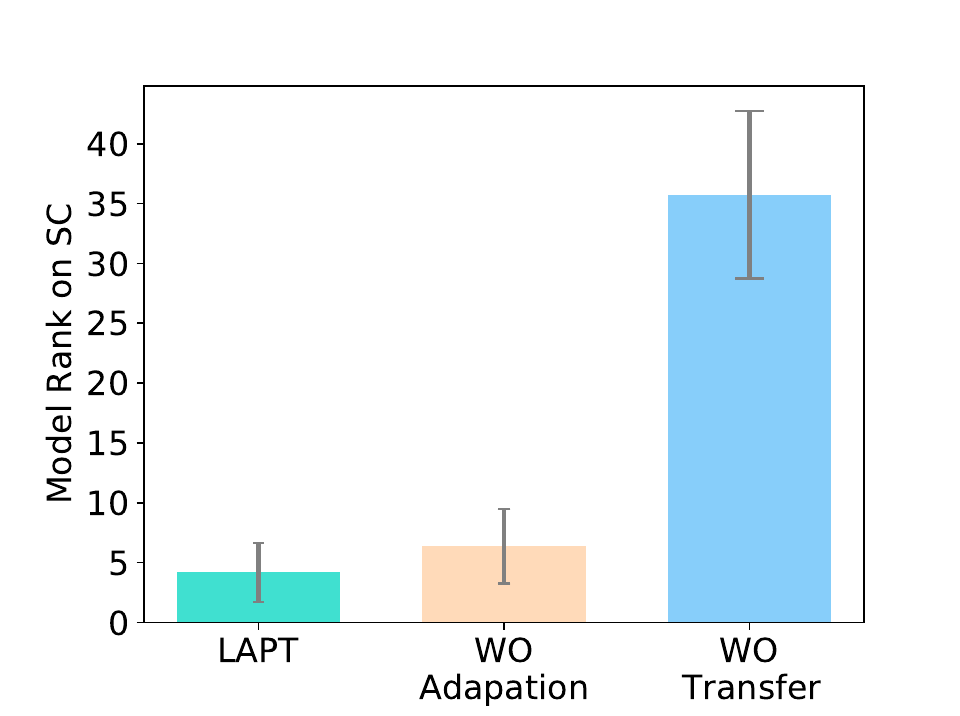}
    \end{minipage}
}

\subfigure[Roo]{
    \begin{minipage}[t]{0.215\textwidth}
    \includegraphics[scale=0.275]{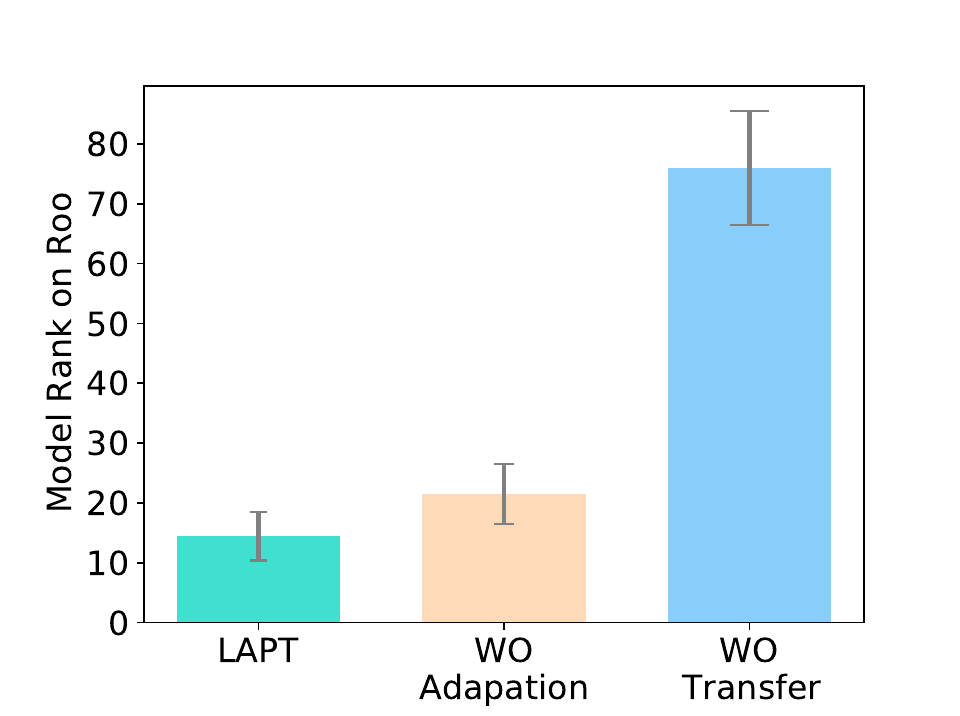}
    \end{minipage}
}
\subfigure[Auto]{
    \begin{minipage}[t]{0.215\textwidth}
    \includegraphics[scale=0.275]{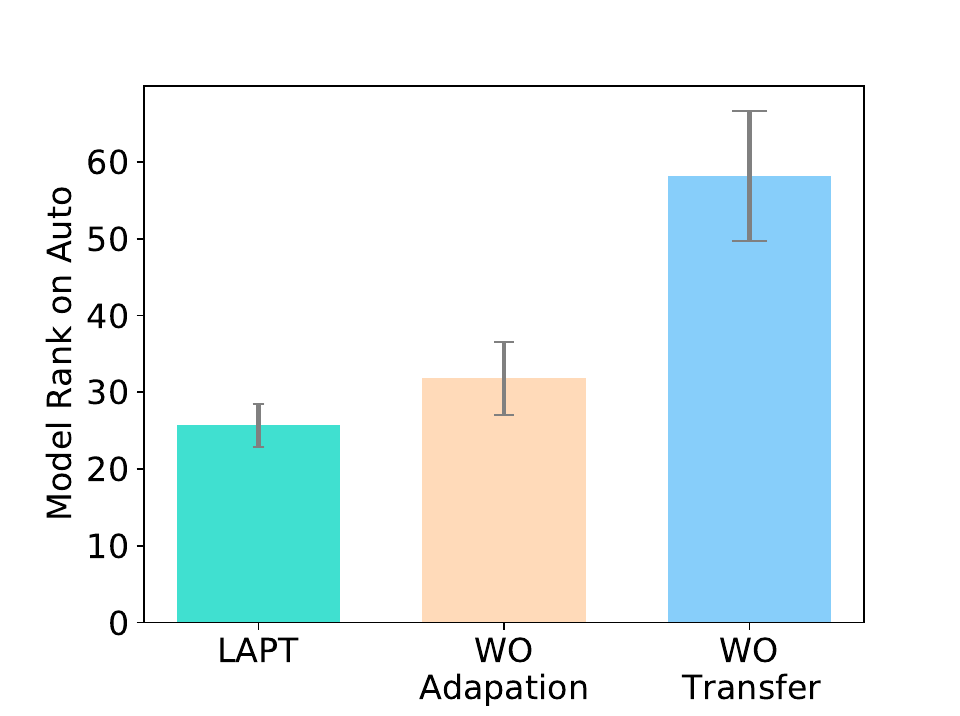}
    \end{minipage}
}

\subfigure[Nor]{
    \begin{minipage}[t]{0.215\textwidth}
    \includegraphics[scale=0.275]{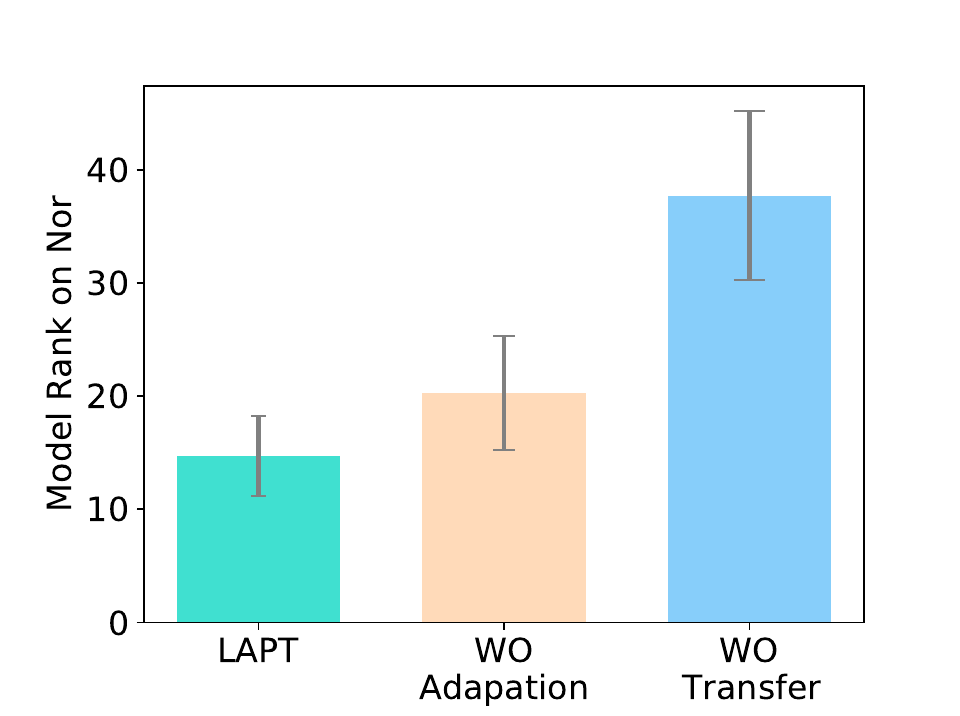}
    \end{minipage}
}
\subfigure[Seg]{
    \begin{minipage}[t]{0.215\textwidth}
    \includegraphics[scale=0.275]{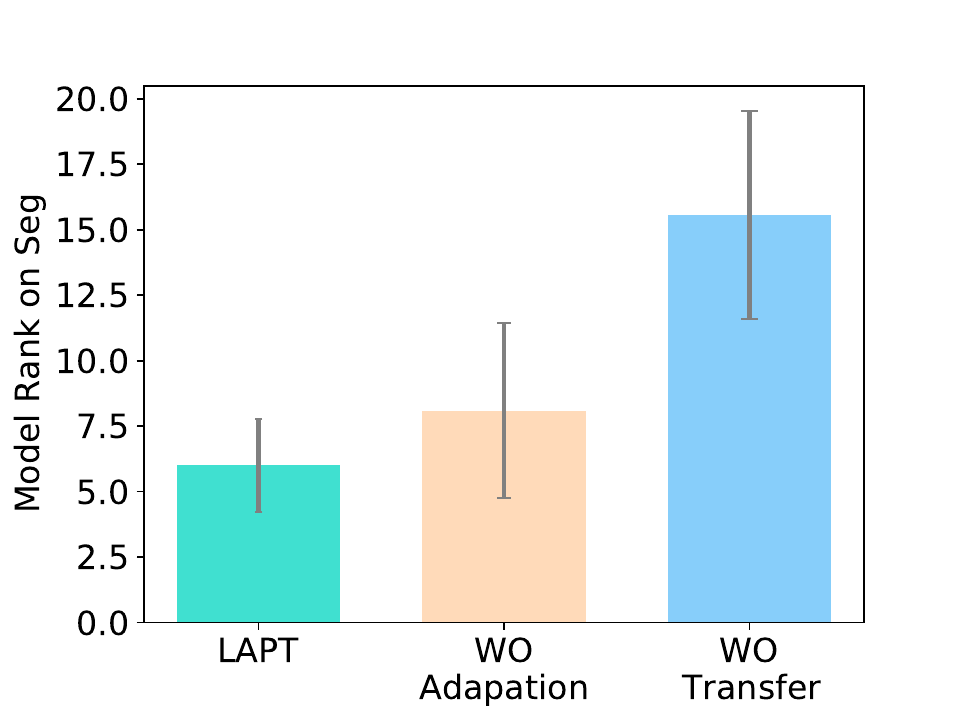}
    \end{minipage}
}
    \caption{Model rank of different LAPT versions on Trans101.}
    \label{ab1}
\end{figure}
In the proposed LAPT framework, two core components are emphasized: knowledge transfer and principle adaptation. To showcase their efficacy, two different verions of LAPT are performed on Trans101. In the first version, the principle adaptation step is omitted, referred to as \textbf{WO Adaptation}. In this scenario, the search space is refined solely based on the design principles derived from previous tasks and never refined during the search process. In the second version, denoted as \textbf{WO Transfer}, architectural knowledge from prior tasks is unavailable, and LAPT refines the search space solely based on the architectures designed for the new task. The model ranks of the discovered architectures via these two versions are illustrated in Fig.~\ref{ab1}. These results demonstrate that knowledge transfer closely aligns with enhanced search performance. Without employing the transfer strategy, identifying top-ranking architectures becomes challenging. Furthermore, the principle adaptation strategy is more related to the convergence of the search method (leading to the lower variance of model ranks).

\section{Conclusion}
This paper introduces a novel idea of design principle transfer in NAS, aiming to improve the architecture search efficiency for new tasks. To this end, an LLM-assist framework is proposed to help learn the design principles from established architectures and transfer them well to the new NAS tasks. Experiments on different search spaces have demonstrated that the quality of the refined search spaces is higher than the original ones. Even the vanilla search method can achieve SOTA results when searching in such space. However, in this work, these tasks need to share the high similarity and the same search space. Knowledge transfer across different search spaces and domains should be explored in the future.
\section{Acknowledgments}
This work was supported in part by National Key R\&D Program of China (2022YFC3801700); in part by the Research Grants Council of the Hong Kong SAR (Grant No. PolyU11211521, PolyU15218622, PolyU15215623, and C5052-23G), and the National Natural Science Foundation of China (Grant No. U21A20512).
\bibliography{aaai22}

\begin{thebibliography}{47}
\providecommand{\natexlab}[1]{#1}

\bibitem[{Baker et~al.(2016)Baker, Gupta, Naik, and Raskar}]{Bowen2016}
Baker, B.; Gupta, O.; Naik, N.; and Raskar, R. 2016.
\newblock Designing neural network architectures using reinforcement learning.
\newblock \emph{arXiv preprint arXiv:1611.02167}.

\bibitem[{Chen, Dohan, and So(2024)}]{chen2024evoprompting}
Chen, A.; Dohan, D.; and So, D. 2024.
\newblock Evoprompting: Language models for code-level neural architecture search.
\newblock \emph{Advances in Neural Information Processing Systems}, 36.

\bibitem[{Chen et~al.(2024)Chen, Xu, Li, Han, Wang, Li, and Hui}]{chen2024large}
Chen, L.; Xu, F.; Li, N.; Han, Z.; Wang, M.; Li, Y.; and Hui, P. 2024.
\newblock Large Language Model-driven Meta-structure Discovery in Heterogeneous Information Network.
\newblock \emph{arXiv preprint arXiv:2402.11518}.

\bibitem[{Chu et~al.(2020)Chu, Zhou, Zhang, and Li}]{chu2020fair}
Chu, X.; Zhou, T.; Zhang, B.; and Li, J. 2020.
\newblock Fair darts: Eliminating unfair advantages in differentiable architecture search.
\newblock In \emph{European conference on computer vision}, 465--480. Springer.

\bibitem[{Cowen-Rivers et~al.(2022)Cowen-Rivers, Lyu, Tutunov, Wang, Grosnit, Griffiths, Maraval, Jianye, Wang, Peters et~al.}]{cowen2022hebo}
Cowen-Rivers, A.~I.; Lyu, W.; Tutunov, R.; Wang, Z.; Grosnit, A.; Griffiths, R.~R.; Maraval, A.~M.; Jianye, H.; Wang, J.; Peters, J.; et~al. 2022.
\newblock Hebo: Pushing the limits of sample-efficient hyper-parameter optimisation.
\newblock \emph{Journal of Artificial Intelligence Research}, 74: 1269--1349.

\bibitem[{Dong et~al.(2023)Dong, Gao, Wang, Yang, and Zhang}]{dong2023heterogeneous}
Dong, H.; Gao, Y.; Wang, H.; Yang, H.; and Zhang, P. 2023.
\newblock Heterogeneous Graph Neural Architecture Search with GPT-4.
\newblock \emph{arXiv preprint arXiv:2312.08680}.

\bibitem[{Dong et~al.(2022)Dong, Hou, Feng, Tang, Tan, and Ong}]{dong2022cell}
Dong, J.; Hou, B.; Feng, L.; Tang, H.; Tan, K.~C.; and Ong, Y.-S. 2022.
\newblock A cell-based fast memetic algorithm for automated convolutional neural architecture design.
\newblock \emph{IEEE Transactions on Neural Networks and Learning Systems}.
\newblock Early access, DOI: 10.1109/TNNLS.2022.3155230.

\bibitem[{Dong and Yang(2020)}]{dong2020bench}
Dong, X.; and Yang, Y. 2020.
\newblock Nas-bench-201: Extending the scope of reproducible neural architecture search.
\newblock \emph{arXiv preprint arXiv:2001.00326}.

\bibitem[{Duan et~al.(2021)Duan, Chen, Xu, Chen, Liang, Zhang, and Li}]{duan2021transnas}
Duan, Y.; Chen, X.; Xu, H.; Chen, Z.; Liang, X.; Zhang, T.; and Li, Z. 2021.
\newblock Transnas-bench-101: Improving transferability and generalizability of cross-task neural architecture search.
\newblock In \emph{Proceedings of the IEEE/CVF Conference on Computer Vision and Pattern Recognition}, 5251--5260.

\bibitem[{Elsken et~al.(2020)Elsken, Staffler, Metzen, and Hutter}]{elsken2020meta}
Elsken, T.; Staffler, B.; Metzen, J.~H.; and Hutter, F. 2020.
\newblock Meta-learning of neural architectures for few-shot learning.
\newblock In \emph{Proceedings of the IEEE/CVF Conference on Computer Vision and Pattern Recognition}, 12365--12375.

\bibitem[{He et~al.(2016)He, Zhang, Ren, and Sun}]{he2016deep}
He, K.; Zhang, X.; Ren, S.; and Sun, J. 2016.
\newblock Deep residual learning for image recognition.
\newblock In \emph{Proceedings of the IEEE Conference on Computer Vision and Pattern Recognition}, 770--778.

\bibitem[{Huang et~al.(2017)Huang, Liu, Van Der~Maaten, and Weinberger}]{huang2017densely}
Huang, G.; Liu, Z.; Van Der~Maaten, L.; and Weinberger, K.~Q. 2017.
\newblock Densely connected convolutional networks.
\newblock In \emph{Proceedings of the IEEE Conference on Computer Vision and Pattern Recognition}, 4700--4708.

\bibitem[{Huang et~al.(2022)Huang, Huang, Li, Chen, Xu, Li, and Liang}]{huang2022arch}
Huang, M.; Huang, Z.; Li, C.; Chen, X.; Xu, H.; Li, Z.; and Liang, X. 2022.
\newblock Arch-Graph: Acyclic Architecture Relation Predictor for Task-Transferable Neural Architecture Search.
\newblock In \emph{Proceedings of the IEEE/CVF Conference on Computer Vision and Pattern Recognition}, 11881--11891.

\bibitem[{Imani, Du, and Shrivastava(2023)}]{imani2023mathprompter}
Imani, S.; Du, L.; and Shrivastava, H. 2023.
\newblock Mathprompter: Mathematical reasoning using large language models.
\newblock \emph{arXiv preprint arXiv:2303.05398}.

\bibitem[{Jawahar et~al.(2023)Jawahar, Abdul-Mageed, Lakshmanan, and Ding}]{jawahar2023llm}
Jawahar, G.; Abdul-Mageed, M.; Lakshmanan, L.~V.; and Ding, D. 2023.
\newblock LLM performance predictors are good initializers for architecture search.
\newblock \emph{arXiv preprint arXiv:2310.16712}.

\bibitem[{Lee and Hyung(2021)}]{lee2021rapid}
Lee, H.; and Hyung, E. 2021.
\newblock Rapid neural architecture search by learning to generate graphs from datasets.
\newblock \emph{arXiv preprint arXiv:2107.00860}.

\bibitem[{Li et~al.(2021)Li, Zhan, Tan, and Zhang}]{li2021meta}
Li, J.-Y.; Zhan, Z.-H.; Tan, K.~C.; and Zhang, J. 2021.
\newblock A meta-knowledge transfer-based differential evolution for multitask optimization.
\newblock \emph{IEEE Transactions on Evolutionary Computation}, 26(4): 719--734.

\bibitem[{Liu et~al.(2024)Liu, Xialiang, Yuan, Lin, Luo, Wang, Lu, and Zhang}]{liu2024evolution}
Liu, F.; Xialiang, T.; Yuan, M.; Lin, X.; Luo, F.; Wang, Z.; Lu, Z.; and Zhang, Q. 2024.
\newblock Evolution of Heuristics: Towards Efficient Automatic Algorithm Design Using Large Language Model.
\newblock In \emph{Forty-first International Conference on Machine Learning}.

\bibitem[{Liu, Simonyan, and Yang(2018)}]{Hanxiao2018}
Liu, H.; Simonyan, K.; and Yang, Y. 2018.
\newblock Darts: Differentiable architecture search.
\newblock \emph{arXiv preprint arXiv:1806.09055}.

\bibitem[{Lu et~al.(2021)Lu, Sreekumar, Goodman, Banzhaf, Deb, and Boddeti}]{lu2021neural}
Lu, Z.; Sreekumar, G.; Goodman, E.; Banzhaf, W.; Deb, K.; and Boddeti, V.~N. 2021.
\newblock Neural architecture transfer.
\newblock \emph{IEEE Transactions on Pattern Analysis and Machine Intelligence}, 43(9): 2971--2989.

\bibitem[{Lu et~al.(2020)Lu, Whalen, Dhebar, Deb, Goodman, Banzhaf, and Boddeti}]{lu2020multiobjective}
Lu, Z.; Whalen, I.; Dhebar, Y.; Deb, K.; Goodman, E.~D.; Banzhaf, W.; and Boddeti, V.~N. 2020.
\newblock Multiobjective Evolutionary Design of Deep Convolutional Neural Networks for Image Classification.
\newblock \emph{IEEE Transactions on Evolutionary Computation}, 25(2): 277--291.

\bibitem[{Nasir et~al.(2023)Nasir, Earle, Togelius, James, and Cleghorn}]{nasir2023llmatic}
Nasir, M.~U.; Earle, S.; Togelius, J.; James, S.; and Cleghorn, C. 2023.
\newblock Llmatic: Neural architecture search via large language models and quality-diversity optimization.
\newblock \emph{arXiv preprint arXiv:2306.01102}.

\bibitem[{Qin et~al.(2024)Qin, Hu, Yan, Xiong, Abbasi, and Shi}]{qin2024fl}
Qin, R.; Hu, Y.; Yan, Z.; Xiong, J.; Abbasi, A.; and Shi, Y. 2024.
\newblock FL-NAS: Towards Fairness of NAS for Resource Constrained Devices via Large Language Models.
\newblock \emph{arXiv preprint arXiv:2402.06696}.

\bibitem[{Radosavovic et~al.(2020)Radosavovic, Kosaraju, Girshick, He, and Doll{\'a}r}]{radosavovic2020designing}
Radosavovic, I.; Kosaraju, R.~P.; Girshick, R.; He, K.; and Doll{\'a}r, P. 2020.
\newblock Designing network design spaces.
\newblock In \emph{Proceedings of the IEEE/CVF Conference on Computer Vision and Pattern Recognition}, 10428--10436.

\bibitem[{Real et~al.(2019)Real, Aggarwal, Huang, and Le}]{Real2019}
Real, E.; Aggarwal, A.; Huang, Y.; and Le, Q.~V. 2019.
\newblock Regularized evolution for image classifier architecture search.
\newblock In \emph{Proceedings of the AAAI Conference on Artificial Intelligence}, 4780--4789.

\bibitem[{Shala et~al.(2023)Shala, Elsken, Hutter, and Grabocka}]{shala2022transfer}
Shala, G.; Elsken, T.; Hutter, F.; and Grabocka, J. 2023.
\newblock Transfer NAS with meta-learned bayesian surrogates.
\newblock In \emph{The Eleventh International Conference on Learning Representations}.

\bibitem[{Shen et~al.(2023)Shen, Wang, Lin, Huang, Tang, Sun, and Wang}]{shen2023deepmad}
Shen, X.; Wang, Y.; Lin, M.; Huang, Y.; Tang, H.; Sun, X.; and Wang, Y. 2023.
\newblock Deepmad: Mathematical architecture design for deep convolutional neural network.
\newblock In \emph{Proceedings of the IEEE/CVF Conference on Computer Vision and Pattern Recognition}, 6163--6173.

\bibitem[{Shi et~al.(2020)Shi, Pi, Xu, Li, Kwok, and Zhang}]{shi2020bridging}
Shi, H.; Pi, R.; Xu, H.; Li, Z.; Kwok, J.; and Zhang, T. 2020.
\newblock Bridging the gap between sample-based and one-shot neural architecture search with bonas.
\newblock In \emph{Advances in Neural Information Processing Systems}, 1808--1819.

\bibitem[{Simonyan and Zisserman(2014)}]{simonyan2014very}
Simonyan, K.; and Zisserman, A. 2014.
\newblock Very deep convolutional networks for large-scale image recognition.
\newblock \emph{arXiv preprint arXiv:1409.1556}.

\bibitem[{Sun et~al.(2018)Sun, Xue, Zhang, and Yen}]{sun2018particle}
Sun, Y.; Xue, B.; Zhang, M.; and Yen, G.~G. 2018.
\newblock A particle swarm optimization-based flexible convolutional autoencoder for image classification.
\newblock \emph{IEEE transactions on neural networks and learning systems}.

\bibitem[{Wang et~al.(2023)Wang, Gao, Zheng, Zhang, Chen, and Bu}]{wang2023graph}
Wang, H.; Gao, Y.; Zheng, X.; Zhang, P.; Chen, H.; and Bu, J. 2023.
\newblock Graph neural architecture search with gpt-4.
\newblock \emph{arXiv preprint arXiv:2310.01436}.

\bibitem[{Wang et~al.(2024)Wang, Ke, Liang, and Zhang}]{wang2024mathnas}
Wang, Q.; Ke, J.; Liang, Z.; and Zhang, S. 2024.
\newblock MathNAS: If Blocks Have a Role in Mathematical Architecture Design.
\newblock \emph{Advances in Neural Information Processing Systems}, 36.

\bibitem[{Wu et~al.(2021)Wu, Dai, Chen, Chen, Liu, Yu, Wang, Liu, Chen, and Yuan}]{wu2021stronger}
Wu, J.; Dai, X.; Chen, D.; Chen, Y.; Liu, M.; Yu, Y.; Wang, Z.; Liu, Z.; Chen, M.; and Yuan, L. 2021.
\newblock Stronger nas with weaker predictors.
\newblock In \emph{Advances in Neural Information Processing Systems}, 28904--28918.

\bibitem[{Wu et~al.(2024)Wu, Wu, Wu, Feng, and Tan}]{wu2024evolutionary}
Wu, X.; Wu, S.-h.; Wu, J.; Feng, L.; and Tan, K.~C. 2024.
\newblock Evolutionary Computation in the Era of Large Language Model: Survey and Roadmap.
\newblock \emph{IEEE Transactions on Evolutionary Computation}.
\newblock Early access, DOI: 10.1109/TEVC.2024.3506731.

\bibitem[{Xiao et~al.(2022)Xiao, Wang, Zhu, Zhou, and Lu}]{xiao2022shapley}
Xiao, H.; Wang, Z.; Zhu, Z.; Zhou, J.; and Lu, J. 2022.
\newblock Shapley-NAS: Discovering operation contribution for neural architecture search.
\newblock In \emph{Proceedings of the IEEE/CVF conference on computer vision and pattern recognition}, 11892--11901.

\bibitem[{Xie et~al.(2018)Xie, Zheng, Liu, and Lin}]{xie2018snas}
Xie, S.; Zheng, H.; Liu, C.; and Lin, L. 2018.
\newblock SNAS: stochastic neural architecture search.
\newblock \emph{arXiv preprint arXiv:1812.09926}.

\bibitem[{Xu et~al.(2020)Xu, Xie, Zhang, Chen, Qi, Tian, and Xiong}]{xu2020pcdarts}
Xu, Y.; Xie, L.; Zhang, X.; Chen, X.; Qi, G.-J.; Tian, Q.; and Xiong, H. 2020.
\newblock PC-DARTs: Partial channel connections for memory-efficient differentiable architecture search.
\newblock In \emph{International Conference on Learning Representations}.

\bibitem[{Xue et~al.(2021)Xue, Wang, Yan, Hu, Yang, and Sun}]{xue2021rethinking}
Xue, C.; Wang, X.; Yan, J.; Hu, Y.; Yang, X.; and Sun, K. 2021.
\newblock Rethinking Bi-Level Optimization in Neural Architecture Search: A Gibbs Sampling Perspective.
\newblock In \emph{Proceedings of the AAAI Conference on Artificial Intelligence}, 10551--10559.

\bibitem[{Yang et~al.(2022)Yang, Tian, Xiang, Peng, and Zhang}]{yang2022accelerating}
Yang, S.; Tian, Y.; Xiang, X.; Peng, S.; and Zhang, X. 2022.
\newblock Accelerating Evolutionary Neural Architecture Search via Multifidelity Evaluation.
\newblock \emph{IEEE Transactions on Cognitive and Developmental Systems}, 14(4): 1778--1792.

\bibitem[{Yuan et~al.(2022)Yuan, Liu, Tian, and Liu}]{yuan2022visual}
Yuan, J.; Liu, M.; Tian, F.; and Liu, S. 2022.
\newblock Visual analysis of neural architecture spaces for summarizing design principles.
\newblock \emph{IEEE Transactions on Visualization and Computer Graphics}, 29(1): 288--298.

\bibitem[{Zagoruyko and Komodakis(2016)}]{zagoruyko2016wide}
Zagoruyko, S.; and Komodakis, N. 2016.
\newblock Wide residual networks.
\newblock \emph{arXiv preprint arXiv:1605.07146}.

\bibitem[{Zhang et~al.(2023)Zhang, Gong, Wu, Liu, and Zhou}]{zhang2023automl}
Zhang, S.; Gong, C.; Wu, L.; Liu, X.; and Zhou, M. 2023.
\newblock Automl-gpt: Automatic machine learning with gpt.
\newblock \emph{arXiv preprint arXiv:2305.02499}.

\bibitem[{Zheng et~al.(2023)Zheng, Su, You, Wang, Qian, Xu, and Albanie}]{zheng2023can}
Zheng, M.; Su, X.; You, S.; Wang, F.; Qian, C.; Xu, C.; and Albanie, S. 2023.
\newblock Can GPT-4 perform neural architecture search?
\newblock \emph{arXiv preprint arXiv:2304.10970}.

\bibitem[{Zhou et~al.(2021{\natexlab{a}})Zhou, Qin, Sun, and Tan}]{zhou2021survey2}
Zhou, X.; Qin, A.; Sun, Y.; and Tan, K.~C. 2021{\natexlab{a}}.
\newblock A Survey of Advances in Evolutionary Neural Architecture Search.
\newblock In \emph{2021 IEEE Congress on Evolutionary Computation (CEC)}, 950--957.

\bibitem[{Zhou et~al.(2021{\natexlab{b}})Zhou, Qin, Gong, and Tan}]{zhou2021survey}
Zhou, X.; Qin, A.~K.; Gong, M.; and Tan, K.~C. 2021{\natexlab{b}}.
\newblock A survey on evolutionary construction of deep neural networks.
\newblock \emph{IEEE Transactions on Evolutionary Computation}, 25(5): 894--912.

\bibitem[{Zhou et~al.(2023)Zhou, Wang, Feng, Liu, Wong, and Tan}]{zhou2023towards}
Zhou, X.; Wang, Z.; Feng, L.; Liu, S.; Wong, K.-C.; and Tan, K.~C. 2023.
\newblock Towards Evolutionary Multi-Task Convolutional Neural Architecture Search.
\newblock \emph{IEEE Transactions on Evolutionary Computation}, 28(3): 682--695.

\bibitem[{Zoph and Le(2016)}]{Barret2016}
Zoph, B.; and Le, Q.~V. 2016.
\newblock Neural architecture search with reinforcement learning.
\newblock \emph{arXiv preprint arXiv:1611.01578}.

\end{thebibliography}

\section{Appendix I Additional Related Work} 
\subsection{Neural architecture search}
Neural architecture search (NAS) has made significant strides in the automatic construction of deep neural network (DNN) architectures~\cite{zhou2021survey}. NAS
typically follows an iterative framework: a search method is utilized to discover promising architectures within a predefined search space. Then, through a tailored performance evaluation system, evaluations of these architectures serve as feedback to guide the search toward better-performing architectures.

Three primary NAS methods have been introduced in existing works, i.e., reinforcement learning-based methods (denoted as RLNAS)~\cite{Bowen2016, Barret2016}, gradient-based NAS methods (denoted as GDNAS)~\cite{Hanxiao2018}, and evolutionary algorithm-based methods (denoted as ENAS)~\cite{sun2018particle}. RLNAS trains an agent to build an architecture layer by layer following a decision-making process. In ENAS, each individual corresponds to an architecture, and the population consisting of these individuals evolves via genetic operations to discover the optimal architecture. During the search process of both RLNAS and ENAS, massive architectures are required to be trained and evaluated, resulting in expensive time cost. To alleviate this challenge, GDNAS methods reformulate NAS as a continuous optimization problem and employ gradient-based methods to drive the search process. However, due to the reliance on gradient information, GDNAS methods often incur high computational memory demands and local optimal neural architectures.
\begin{figure*}[t]
    \centering
    \includegraphics[width=0.88\linewidth]{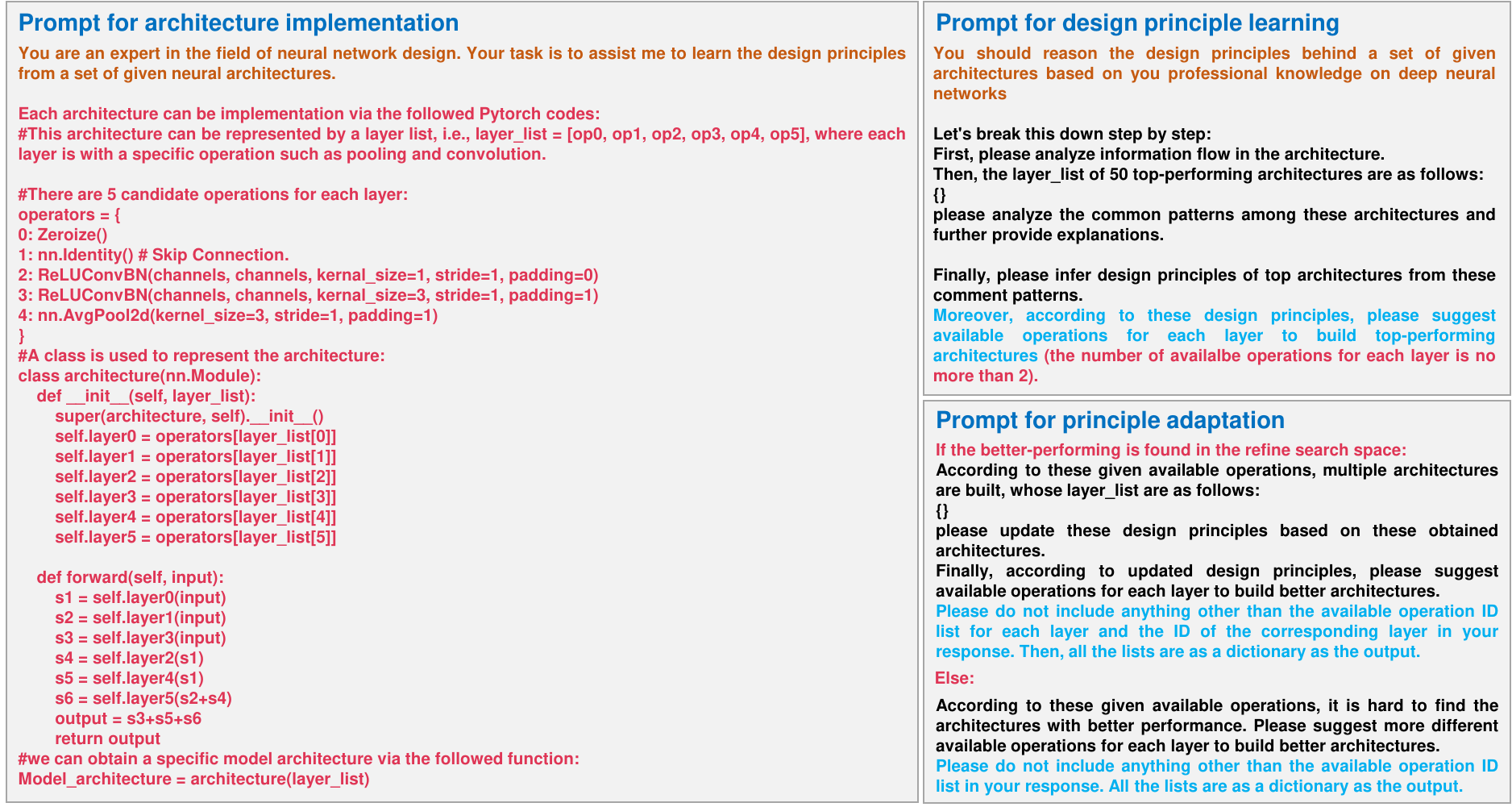}
    \caption{Prompts used to drive LLMs to learn and adapt the design principles from architectures in NAS201.}
    \label{prompt_201}
\end{figure*}

 \begin{figure*}[h!]
    \centering
    \includegraphics[width=0.88\linewidth]{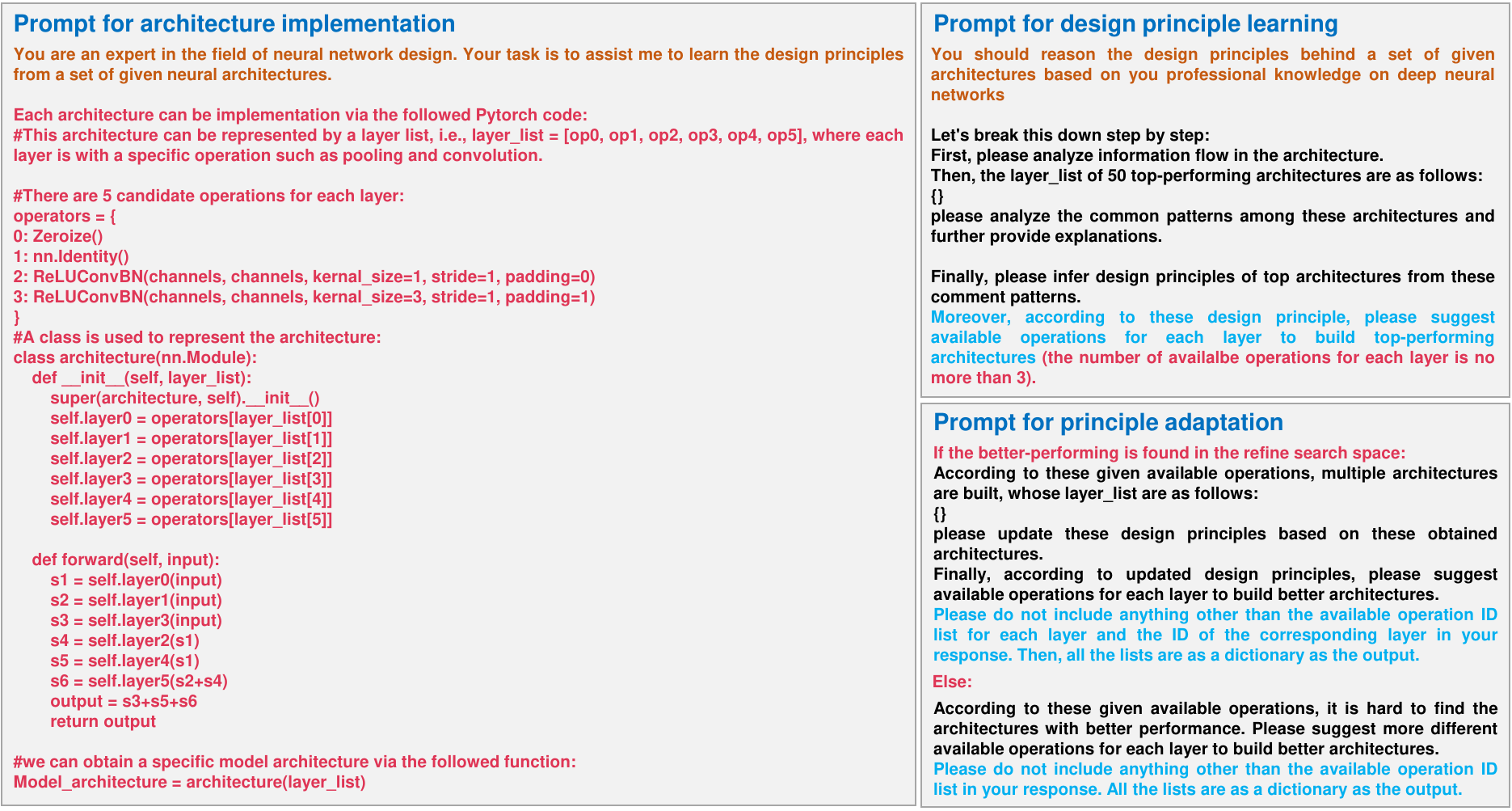}
    \caption{Prompts used to drive LLMs to learn and adapt the design principles from architectures in Trans101.}
    \label{prompt_101}
\end{figure*}

 \begin{figure*}[t!]
    \centering
    \includegraphics[width=0.88\linewidth]{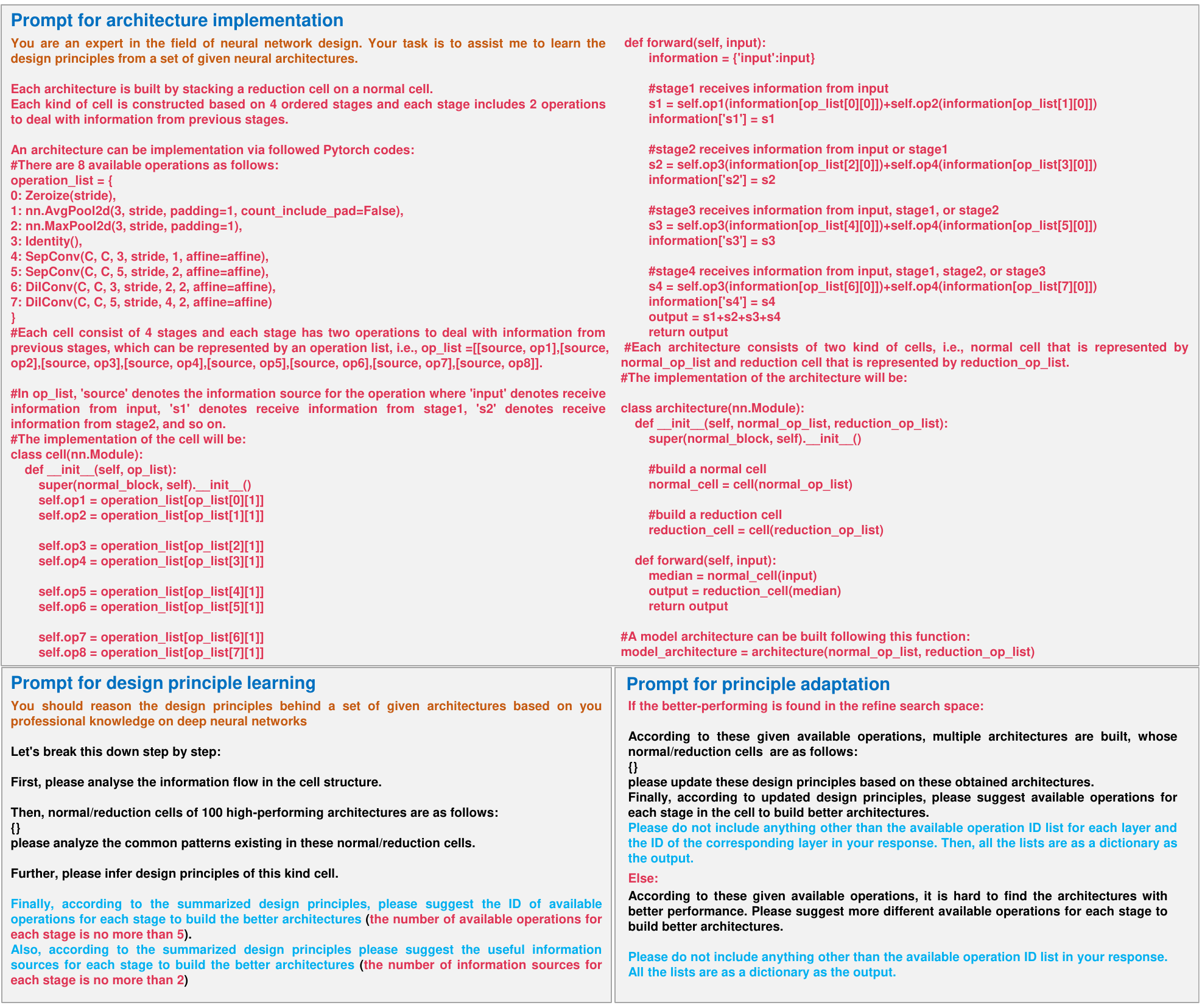}
    \caption{Prompts used to drive LLMs to learn and adapt the design principles from architectures in DARTs.}
    \label{prompt_darts}
\end{figure*}

\subsection{Design principles of DNNs}
Recently, \textit{design in a principled way} has been proposed to reduce the search space in NAS~\cite{radosavovic2020designing,shen2023deepmad}. In these approaches, design principles, which describe how various structural components (such as pooling, convolution, and their interconnections) impact architecture performance, are leveraged to optimize the search space. Specifically, architectures containing less critical components are pruned from the search space according to these principles, resulting in a more refined subspace. This refined subspace contains a higher proportion of high-performing architectures than the original search space \cite{radosavovic2020designing}. Exploration within this refined search space significantly enhances the search effectiveness and efficiency. For instance, by incorporating principles like uniform stage depth and a non-decreasing number of channels, a CNN search space can be condensed from 10$^{10}$ to 10$^{2}$, enabling the discovery of top-performing architectures within seconds~\cite{wang2024mathnas}. However, learning design principles heavily relies on expert knowledge. Additionally, given the intricate and diverse nature of DNN architectures, it is challenging for experts to learn general design principles. These problems restrict the practical application of design principles in real-world scenarios.

\section{Appendix II Prompt Engineering}
In this work, we design specific prompts to help LLMs reason the general design principles from given architectures, translate these principles into architectural parameters, and adapt these principles to specific tasks. These prompts consist of one or several parts:
\begin{enumerate}
    \item {\color{brown} Task description}: It informs LLMs of the problem that is to be solved.
    \item {\color{black} Strategy}: It informs LLMs of the steps that should be followed to solve the given problem.
    \item {\color{magenta} Expected output}: It informs LLMs of the type of information that they should output.
    \item {\color{cyan} Note}: It provides additional instructions for LLMs to improve their efficiency.
\end{enumerate}

Prompts used in different spaces share a similar template but with a few differences among their contents. Details of these prompts are shown in Figure~\ref{prompt_201},~\ref{prompt_101}, and~\ref{prompt_darts}.

\begin{figure*}[h!]
\centering
\subfigure[Obj]{
    \begin{minipage}[t]{0.3\textwidth}
    \includegraphics[scale=0.35]{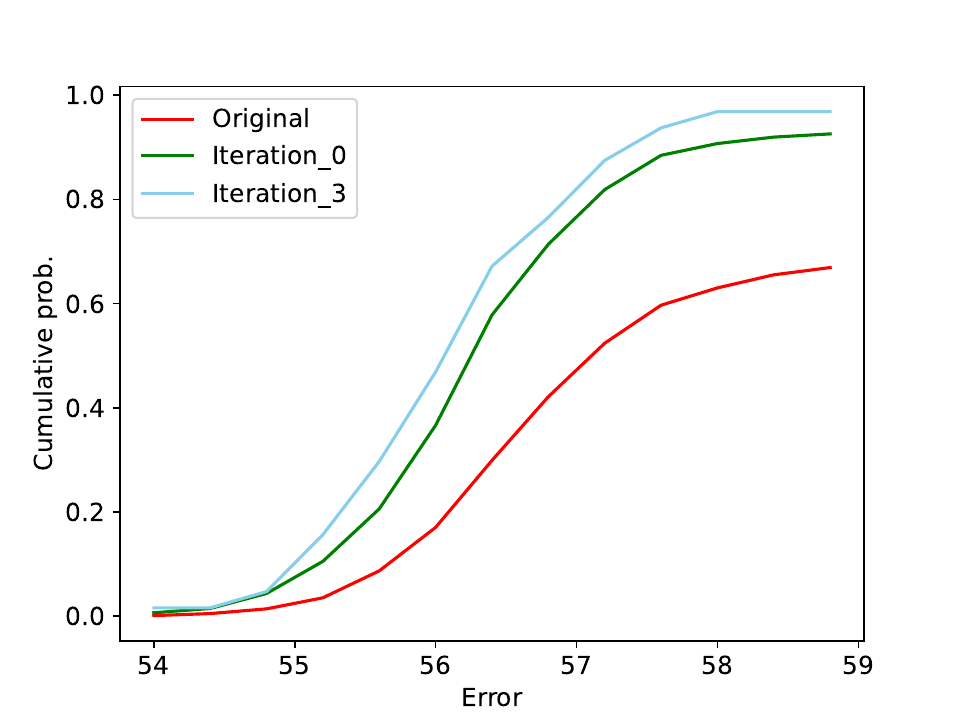}
    \end{minipage}
}
\subfigure[SC]{
    \begin{minipage}[t]{0.3\textwidth}
    \includegraphics[scale=0.35]{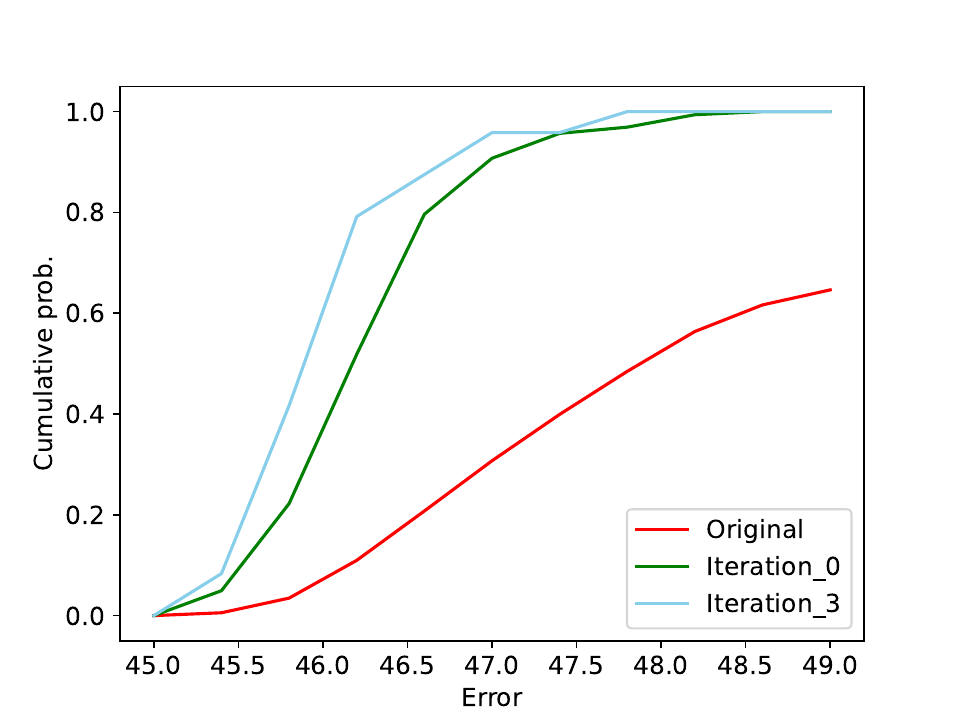}
    \end{minipage}
}
\subfigure[Roo]{
    \begin{minipage}[t]{0.3\textwidth}
    \includegraphics[scale=0.35]{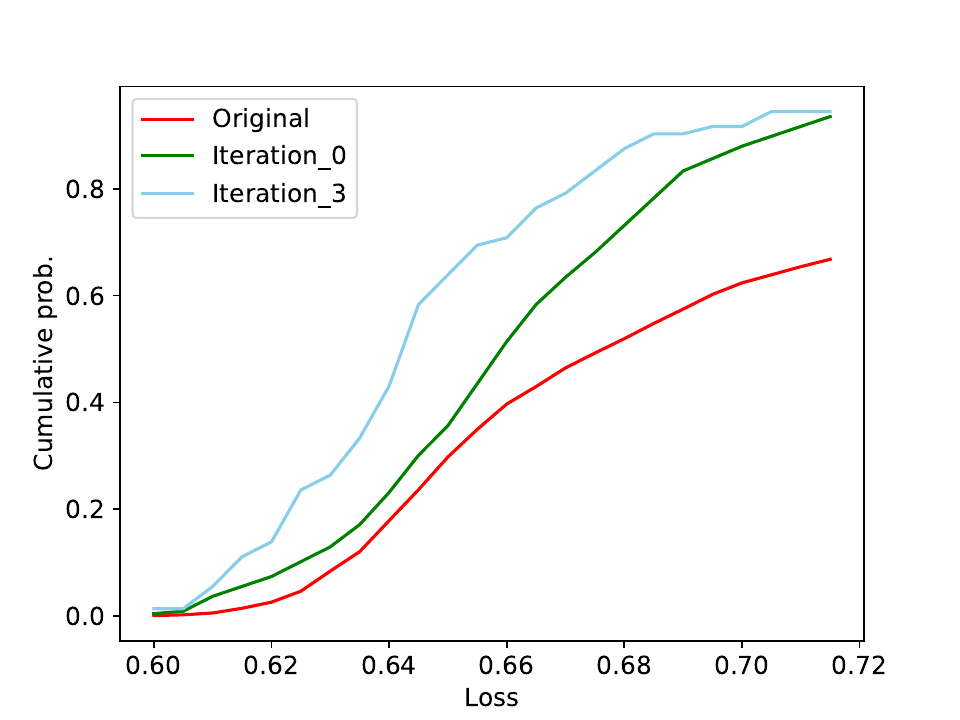}
    \end{minipage}
}

\subfigure[Auto]{
    \begin{minipage}[t]{0.3\textwidth}
    \includegraphics[scale=0.35]{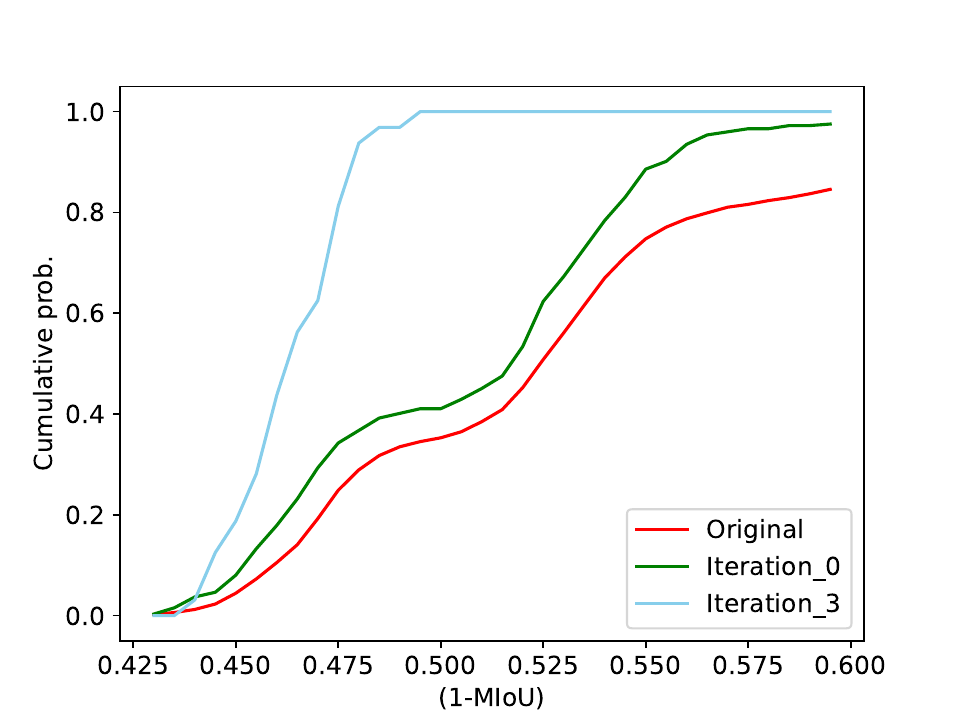}
    \end{minipage}
}
\subfigure[Nor]{
    \begin{minipage}[t]{0.3\textwidth}
    \includegraphics[scale=0.35]{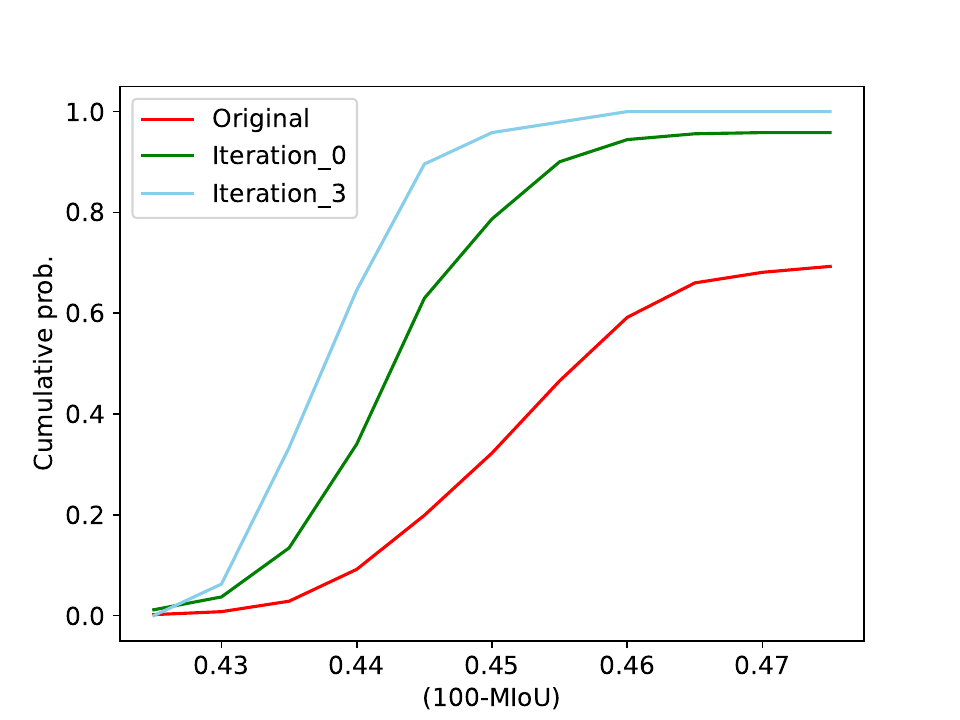}
    \end{minipage}
}
\subfigure[Seg]{
    \begin{minipage}[t]{0.3\textwidth}
    \includegraphics[scale=0.35]{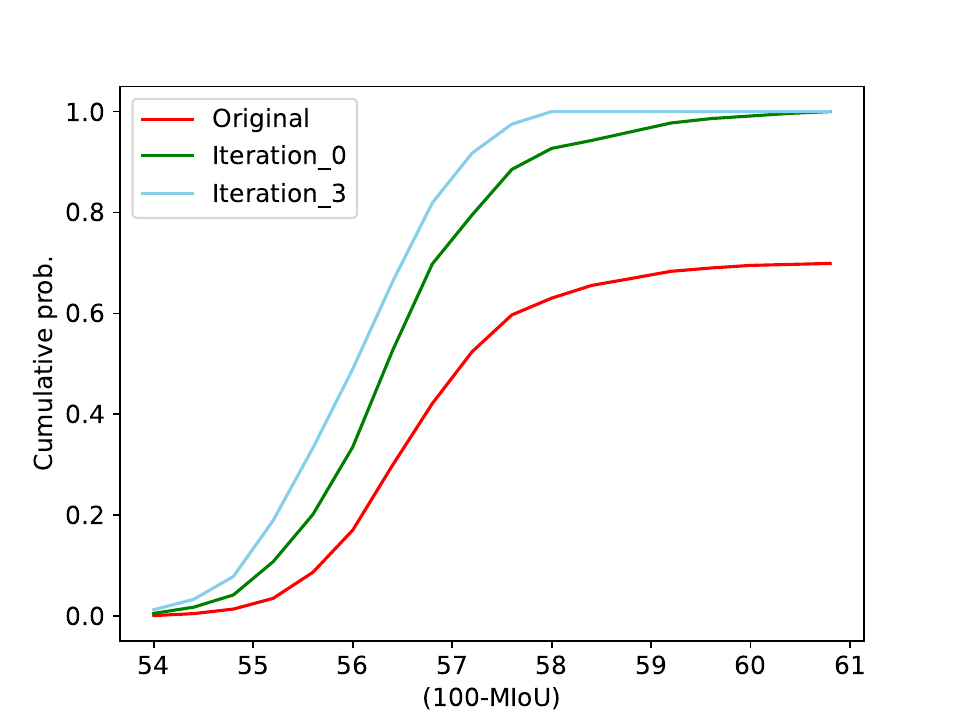}
    \end{minipage}
}
    \caption{Quality of search spaces on Trans101.}
    \label{search space quality}
\end{figure*}
\section{Appendix III Quality Analysis of Search Space}
In this work, the proposed framework LAPT optimizes the search space by pruning the unpromising architectures from it. In this way, a refined search space with a higher proportion of well-performing architectures can be built. To demonstrate the effectiveness of this method, we measure the quality of these search spaces from Transfer101 and present these results in this section. 

Specifically, the error empirical distribution function (eEDF) is used to measure the quality of the search space, which is shown in equation (1):
\begin{equation}
F(e) = \frac{1}{n}\sum_{i=1}^n{1[e_i\leq e]}.
\end{equation}
$F(e)$ gives the fraction of models with errors less than $e$, and $n$ is the total number of architectures in the search space or the subspace. We conduct the proposed LAPT on Trans101 to solve the 6 computer vision tasks (i.e., Object Classification (Obj), Scene Classification (SC), Room Layout (Roo), AutoEncoder (Auto), Surface Normal (Nor), Semantic Segmentation (Seg)) based on the architectural knowledge from the prior task Jigsaw. For each task, we measure the quality of the entire search space (termed original), the refined search space based on the general design principles (termed iteration\_0), and the task-specific search space built based on the adapted design principles (termed iteration\_3). The measurements are shown in Figure~\ref{search space quality}.

From these results, these task-specific search spaces have the highest proportion of well-performing architectures. Then, the ones refined based on the general design principles are also of higher quality than the original one. These results demonstrate that: (1) the proposed LLM-based strategy can well learn the useful general design principles that can be used to identify the unpromising/promising architectures; (2) the design principle transfer paradigm is effective for the search space refinement in new NAS tasks; (3) our principle adaptation is effective in building the task-specific search space which is more effective for the specific NAS task.  
\section{Appendix IV General Design Principles}
In this work, we drive the pre-trained LLMs to learn design principles for different search spaces from established architectures. These general design principles can be reused to prune the search space, leading to a refined search space for the new NAS tasks. The design principles of the architecture in NAS201 have been shown in the main paper. In this part, we present the design principles of the search spaces Trans101 and DARTs, respectively. 

\begin{figure*}[h]
    \centering
    \includegraphics[width=1\linewidth]{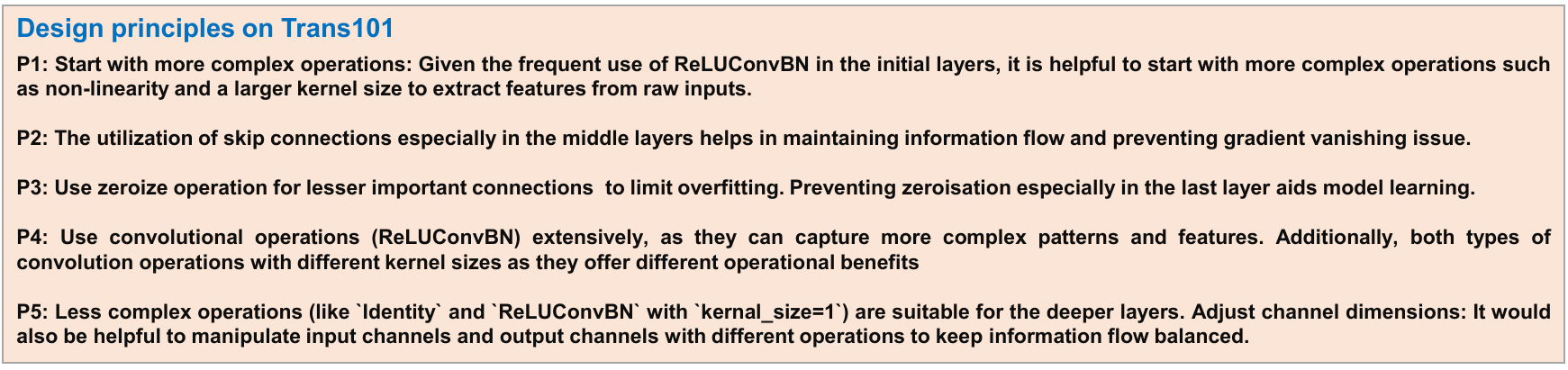}
    \caption{General design principles on Trans101.}
    \label{principles_101}
\end{figure*}

Firstly, the design principles for Trans101 are shown in Figure~\ref{principles_101}. In this search space, these principles mainly describe the influence of different operators (e.g., pooling layer, skip, and convolution), because the number of layers and connections in these architectures are fixed. These principles first point out convolution layers should be used in the early layers of the architecture to extract feature information, and the skip operator should be used in the middle layers to prevent gradient vanishing issues. This is identical to the expert knowledge on convolution neural networks (CNNs). Additionally, the usage of other layers is also presented.

\begin{figure*}[h]
    \centering
    \includegraphics[width=1\linewidth]{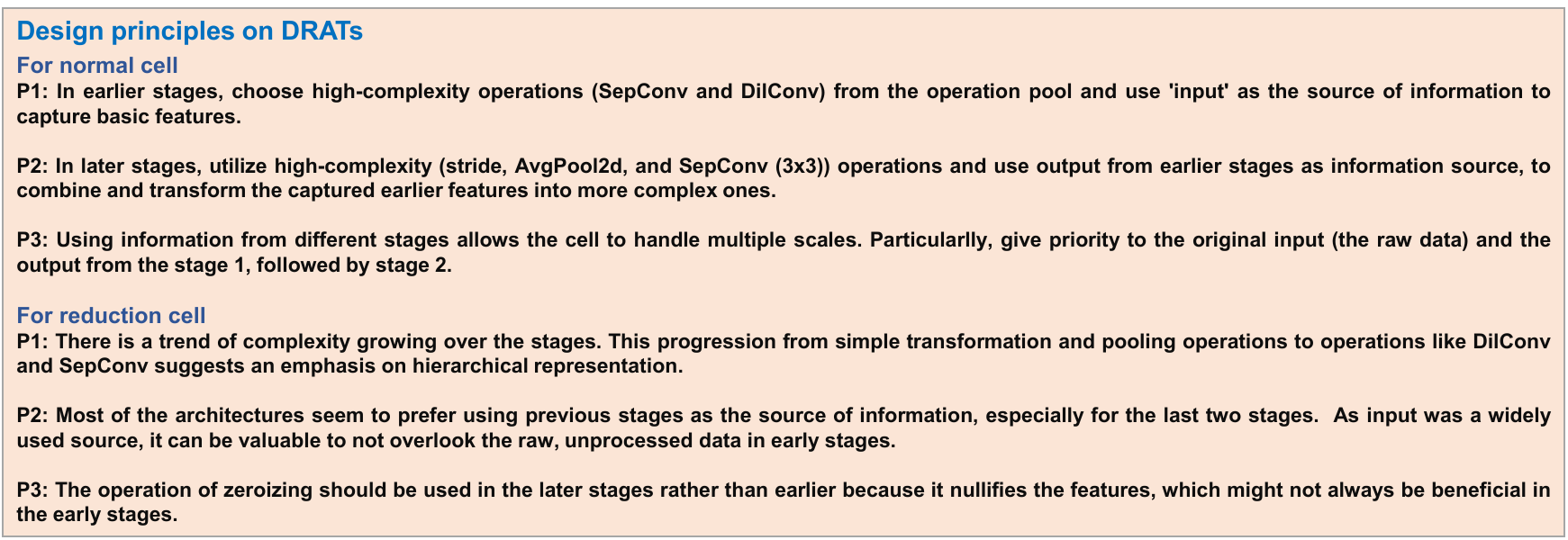}
    \caption{General design principles on Trans101.}
    \label{principles_DARTs}
\end{figure*}

\begin{table*}[h!]
  \centering
  \caption{Mode rank of the found architecture on each task.}
  \scriptsize
  \begin{tabular}{llllllll}
    \hline
    \hline
    {\bf LLM}&{\bf Obj}&{\bf SC}&{\bf Roo}&{\bf Auto}&{\bf Nor}&{\bf Seg}&{\bf Ave.}\\
    \hline
    \vspace{0mm}
    GPT-3.5&40.3$\pm$8.1&16.7$\pm$3.9&61.0$\pm$8.3 &28.3$\pm$5.6&51.2$\pm$10.4&61.9$\pm$11.2&43.2\\
    GPT-4&9.0$\pm$3.0 &4.2$\pm$2.5 &14.4$\pm$4.1 &25.7$\pm$2.8&14.7$\pm$3.5&6.0$\pm$1.8&12.3\\
    GPT-4o&9.9$\pm$2.9 &8.8$\pm$4.0 &9.9$\pm$2.8&23.3$\pm$3.0 &9.3$\pm$2.3&11.4$\pm$3.0&12.1\\
    \hline
    \hline
  \end{tabular}
  \label{Trans101}
\end{table*}

Secondly, the design principles for DARTs are shown in Figure~\ref{principles_DARTs}. In this search space, these principles mainly describe the available operators in each stage and its information source. Because two kinds of cells should be designed, i.e., the normal cell and the reduction cell, we present the design principles of them separately. Specifically, for the normal cell, the complexity of the operators increases with the cell's depth, where the high-complexity operators such as convolution are used to extract feature information in the early stage and the low-complexity operators such as skip and pooling are used in the later stage to prevent gradient vanishing. However, the reduction cell is designed in a different way from the normal one. In the reduction cell's early stage, the low-complexity operators such as pooling are used for the feature size reduction, and high-complexity operators are used in the later stages for feature extraction. The reason is that these cells work in different mechanisms. As for the information sources, all the stages intend to receive information from the previous one to build a deep structure. More details can are in Figure~\ref{principles_DARTs}.

\section{Appendix V Effectiveness of LLMs}
In our proposed LART framework, the pre-trained LLMs play an important role in design principle learning, design principle adaptation, and design principle translation. This part further discusses the effectiveness of LLMs on the performance of LAPT. To this end, we conduct LLMs with different pre-trained LLMs on Trans101, and the model ranks of architectures found for each task are shown in Table I.

From these results, GPT-4 and GPT-4o achieve comparable performance on these tasks, and both of them outperform GPT-3.5 by a large margin. The reason is that GPT-4 and GPT-4o have more powerful language reasoning, understanding, and generation capabilities. It demonstrates that the performance of the proposed LAPT framework is close to the used LLMs. 
\end{document}